\documentclass{article}

\usepackage{microtype}
\usepackage{graphicx}
\usepackage{booktabs} % for professional tables

\usepackage{wrapfig}
\usepackage{multirow}
\usepackage{makecell}
\usepackage[utf8]{inputenc} % allow utf-8 input
\usepackage[T1]{fontenc}    % use 8-bit T1 fonts
\usepackage{hyperref}       % hyperlinks
\usepackage{url}            % simple URL typesetting
\usepackage{booktabs}       % professional-quality tables
\usepackage{amsfonts}       % blackboard math symbols
\usepackage{nicefrac}       % compact symbols for 1/2, etc.
\usepackage{microtype}      % microtypography
\usepackage{xcolor}         % colors
\usepackage{enumitem}
\usepackage{circledsteps}
\usepackage{pifont}
\usepackage{mathtools}
\usepackage{amsmath,amssymb,amsthm}
\usepackage{ulem}
\usepackage{adjustbox}
\usepackage{subcaption}
\usepackage{float}

\newcommand{\WORK}[1]{{SampleAttention}}

\usepackage{bm}       

\usepackage[ruled,vlined]{algorithm2e}

\definecolor{commentcolor}{RGB}{110,154,155}   % define comment color
\definecolor{comment_color}{HTML}{1B8F44}
\definecolor{comment_color_2}{RGB}{64,128,128}
\newcommand{\LineComment}[1]{\vspace*{0.5em}\small\textcolor{comment_color_2}{\textit{\# #1}}}

\usepackage{hyperref}

\usepackage[accepted]{mlsys2025}

\mlsystitlerunning{SampleAttention: Near-Lossless Acceleration of Long Context LLM Inference with Adaptive Structured Sparse Attention}

\begin{document}

\twocolumn[
\mlsystitle{SampleAttention: Near-Lossless Acceleration of Long Context LLM Inference with Adaptive Structured Sparse Attention}

% It is OKAY to include author information, even for blind
% submissions: the style file will automatically remove it for you
% unless you've provided the [accepted] option to the mlsys2025
% package.

% List of affiliations: The first argument should be a (short)
% identifier you will use later to specify author affiliations
% Academic affiliations should list Department, University, City, Region, Country
% Industry affiliations should list Company, City, Region, Country

% You can specify symbols, otherwise they are numbered in order.
% Ideally, you should not use this facility. Affiliations will be numbered
% in order of appearance and this is the preferred way.
\mlsyssetsymbol{equal}{*}

\begin{mlsysauthorlist}
\mlsysauthor{Qianchao Zhu}{equal,pku}
\mlsysauthor{Jiangfei Duan}{equal,cuhk}
\mlsysauthor{Chang Chen}{pku}
\mlsysauthor{Xiuhong Li}{pku}
\mlsysauthor{Siran Liu}{pku}
\mlsysauthor{Guanyu Feng}{zp}
\mlsysauthor{Xin Lv}{zp}
\mlsysauthor{Chuanfu Xiao}{pku}
\mlsysauthor{Dahua Lin}{cuhk}
\mlsysauthor{Chao Yang}{pku}
\end{mlsysauthorlist}

\mlsysaffiliation{pku}{Peking University}
\mlsysaffiliation{cuhk}{The Chinese University of Hong Kong}
\mlsysaffiliation{zp}{Zhipu.AI}

\mlsyscorrespondingauthor{Chao Yang}{chao\_yang@pku.edu.cn}

% You may provide any keywords that you
% find helpful for describing your paper; these are used to populate
% the "keywords" metadata in the PDF but will not be shown in the document
\mlsyskeywords{Machine Learning, MLSys}

\vskip 0.3in

\begin{abstract}
  Large language models (LLMs) now support extremely long context windows, but the quadratic complexity of vanilla attention results in significantly long Time-to-First-Token (TTFT) latency. Exisiting sparse attention approaches employ either static sparse pattern or fixed sparsity ratio to utilize the high attention sparsity, failing to capture the adaptive sparsity ratio and dynamic sparse pattern across attention heads, input contents and model architectures. 
  To balance accuracy and performance efficiently, we introduce a robust indicator for accuracy, Cumulative Residual Attention (CRA), which measures the percentage of attention recall.
  Leveraging this key insight, we present \WORK{}, which employs a novel two-stage query-guided key-value filtering approach to efficiently and dynamically select a minimal set of important column and slash strips to meet a desired CRA threshold, thus maximizing efficiency while preserving accuracy. Comprehensive evaluations show that \WORK{} can establish a new Pareto frontier in the accuracy-efficiency trade-off, and reduces TTFT by up to $5.29\times$ compared with FlashAttention2.
\end{abstract}
]

% \printAffiliationsAndNotice{}  % leave blank if no need to mention equal contribution
\printAffiliationsAndNotice{\mlsysEqualContribution} % otherwise use the standard text.

\section{Introduction}
\begin{figure}[ht]
  \centering
  \includegraphics[width=1.00\linewidth]{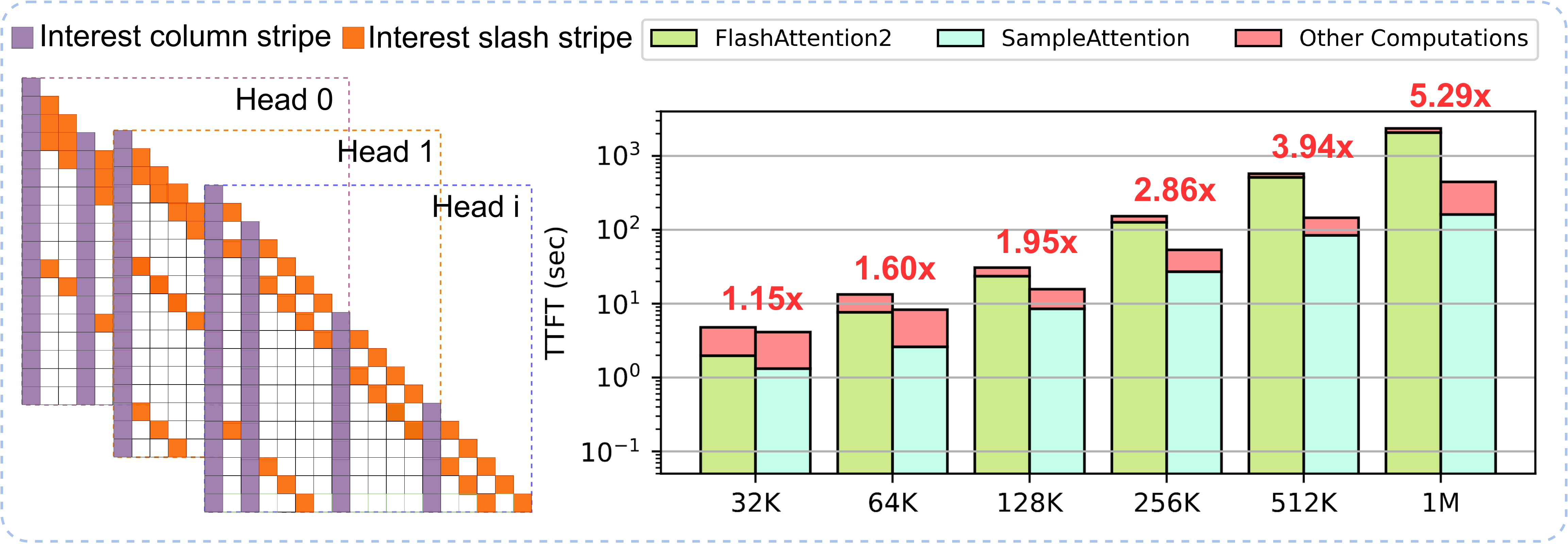}
  \caption{Compared to previous static and dynamic sparse attention methods, \WORK{} captures adaptive structured sparse patterns for each head. It achieves a significant reduction in TTFT compared to FlashAttention2.}
  \label{fig:intro}
  \vspace{-1em}
\end{figure}

Recent advances~\cite{xiong2023effective, liu2023scaling, chen2023longlora, longchat2023, chen2023extending} race to scale the context window of large language models (LLMs)~\cite{gpt-3, vaswani2017attention, touvron2023llama} for more complex applications, including document analysis~\cite{zhang2024benchmarking}, code copilot~\cite{chen2021evaluating, roziere2023code}, and prolonged conversations~\cite{vicuna2023, alpaca}. Popular LLMs like Gemini~\cite{team2023gemini}, Claude~\cite{claude} and Kimi~\cite{kimi} now support context lengths exceeding 1 million tokens. However, the increase in context length makes it challenging to support live interactions due to the quadratic complexity of attention mechanism. As illustrated in Figure~\ref{fig:intro}, the attention computation time increases quadratically with sequence length, quickly dominating the \textit{Time to First Token} (TTFT) latency (i.e. prefill latency). For example, in a 1 million token context, the attention of ChatGLM3-6B~\cite{du2021glm} takes $1555$ seconds, constituting over 90\% of the TTFT when evaluated on an A100 GPU.
 
Prior work has consistently demonstrated that attention scores exhibit high sparsity~\cite{zaheer2020big,kitaev2020reformer,jiang2024minference,li2024snapkv}. This characteristic makes sparse attention a promising approach for reducing prefill latency, as it allows the model to compute attention selectively for only the most important query and key-value tokens rather than the entire sequence. Based on these observations, plenty of approaches propose to approximate the dense attention with static sparse pattern, like LongFormer~\cite{beltagy2020longformer}, BigBird~\cite{zaheer2020big}, LongNet~\cite{ding2023longnet} and StreamingLLM~\cite{xiao2023efficient}. However, these approaches fail to capture the dynamic sparse pattern across heads and inputs~\cite{jiang2024minference,likhosherstov2021expressive} and cannot achieve the same accuracy of full attention. 

Recent work proposes to address this dynamic sparse pattern through runtime attention index selection~\cite{liu2022dynamic,han2023hyperattention,jiang2024minference}. DSA~\cite{liu2022dynamic} approximates attention patterns using low-rank hidden dimensions, but incurs significant computational overhead with long contexts. While HyperAttention~\cite{han2023hyperattention} employs Locality Sensitive Hashing (LSH) to identify important attention scores, its coarse-grained selection approach struggles to maintain model accuracy. MInference~\cite{jiang2024minference} takes a hybrid approach: it predefines several sparse patterns and matches each attention head to its optimal pattern offline to achieve a target sparsification budget, then dynamically searches for sparse indices during the prefill phase. However, this approach's reliance on predefined patterns and fixed budgets fails to capture both the varying sparsity ratios across attention heads and the dynamic sparse patterns that adapt to different input contents.

In this paper, we find effectively exploiting the inherently-high attention sparsity to accelerate prefill computation remains challenging due to two key factors. First, the optimal sparsity ratio varies adaptively across attention heads, input contents, and model architectures, making it necessary to be determined at runtime. Second, attention patterns also varies across heads and contents, often combining typical column and slash patterns. Some attention heads even display intricate combinations of these patterns, further complicating sparse pattern selection. 
These dynamic characteristics create significant challenges for existing methods to achieve an optimal trade-off due to their lack of flexibility. This highlights the need for a more adaptable, runtime-efficient approach to determine both the sparsity ratio and the pattern.
% \red{cause what kind of problem? need some metric? compare with exsiting methods?}

To address these challenges, we propose a novel approach, \WORK{}, which can dynamically determine sparse ratios and patterns at runtime. To select a minimal set of significant column and slash patterns while maintaining accuracy, we introduce a robust metric for evaluating model accuracy called \textit{Cumulative Residual Attention} (\textbf{CRA}), which measures the capability of attention recall. The details of this key insight will be presented in \textit{Section~\ref{sec:motiv}}. 
Based on CRA, we elaborate on a two-stage query-guided key-value filtering method proposed by \WORK{} in \textit{Section~\ref{sec:method}}. This implementation is designed to efficiently identify important columns and slashes at runtime. 
\WORK{} also develops an automated tuning method that uses a small profiling dataset to determine the optimal hyperparameter setting for each model across different length ranges. 
\WORK{} significantly accelerates vanilla attention by reducing both I/O and computation requirements. We also implement hardware-efficient kernels. Notably, \WORK{} aims to reduce the computation overhead of attention, and is orthogonal and can be combined with existing KV cache eviction approaches~\cite{zhang2024h2o,ribar2023sparq,mu2024learning} to further reduce memory consumption.

We evaluate \WORK{} on ChatGLM~\cite{glm2024chatglm}, YI~\cite{young2024yi} and InternLM~\cite{cai2024internlm2} with a suite of popular benchmarks covering various generative tasks across different sequence lengths. Experimental results show that \WORK{} achieves nearly no accuracy loss\footnote{Near-lossless refers to that model accuracy stays above $99\%$ of the baseline according to MLPerf~\cite{reddi2020mlperf}.} for different LLMs, significantly outperforming prior works, and reduces the TTFT by up to $5.29\times$ compared with FlashAttention2.

\section{Background}
\label{sec:bg}

\subsection{LLM Inference}

LLMs are built upon transformer architectures~\cite{vaswani2017attention}, which stacks multiple transformer decoder blocks. Each block consists of an attention layer (Self-Attention) followed by a feed-foward network (MLP). 

The inference process of LLMs operates in two phases: \textit{prefill} and \textit{decoding}. During the prefill phase, the model processes the entire input prompt in parallel and generates the first output token. This phase also generates and stores the Key-Value (KV) cache for each token in the prompt, which will be used in subsequent computations. The decoding phase follows the prefill phase and sequentially generates each new token based on all previous tokens. The model takes one output token as input each time, and leverages the KV cache to generate the subsequent new token autoregressively. The KV cache of the input token will also be saved as the context for subsequent generation.

When processing long input sequences, the computational demands of handling lengthy prompts can result in significant \textit{Time To First Token} (TTFT) latency (i.e. prefill latency), creating a substantial bottleneck for real-world applications. For example, the TTFT of 1 million sequence for ChatGLM-6B~\cite{du2021glm} takes near 30 minutes. This prohibitive latency makes it impractical for applications requiring real-time responses. Therefore, reducing TTFT for long sequences becomes crucial for enabling practical deployment of LLMs in scenarios demanding both long context processing and responsive interaction.

\subsection{Attention Computation}

The attention layer enables the model to weigh the importance of different tokens in the input sequence and dynamically adjust their influence on the output. We start with a regular full attention for one attention head to examine the mechanism, while the following contents can be seamlessly applied to multiple attention heads.

In attention layer, each token in the input sequence is transformed into three vectors: query, key and value tensors. Let $\textbf{Q} \in \mathbb{R}^{S_q \times d}$ and $\textbf{K, V} \in \mathbb{R}^{S_k \times d}$ be the query and key-value tensor of one head, where $S_q, S_k$ is the sequence length respectively, and $d$ is the head dimension. The full attention output $\textbf{O} \in \mathbb{R}^{S_q \times d}$ can be formulated as,
\begin{equation}
\textbf{P} = \texttt{softmax}(\frac{\textbf{QK}^{T}}{\sqrt{d}}) \in [0, 1]^{S_q \times S_k} 
\end{equation}
\begin{equation}
\textbf{O}= \textbf{PV} \in \mathbb{R}^{S_q \times d} 
\end{equation}

where \texttt{softmax} is applied in row-wise, and $\textbf{P}$ is the attention score. 

This attention mechanism poses a fundamental challenge during long-context inference: both the memory footprint of $\textbf{P}$ and the computational complexity scale quadratically with sequence length. While FlashAttention~\cite{dao2022flashattention} effectively addresses the memory bottleneck through online softmax computation, the quadratic computational complexity remains unresolved, resulting in substantial response delays. As previously discussed, attention computation can constitute over $90\%$ of the total TTFT latency, making its optimization crucial for achieving practical long-context inference.

\begin{figure}[t]
  \centering
  \vspace{-0.5em}
  \includegraphics[width=0.85\linewidth]{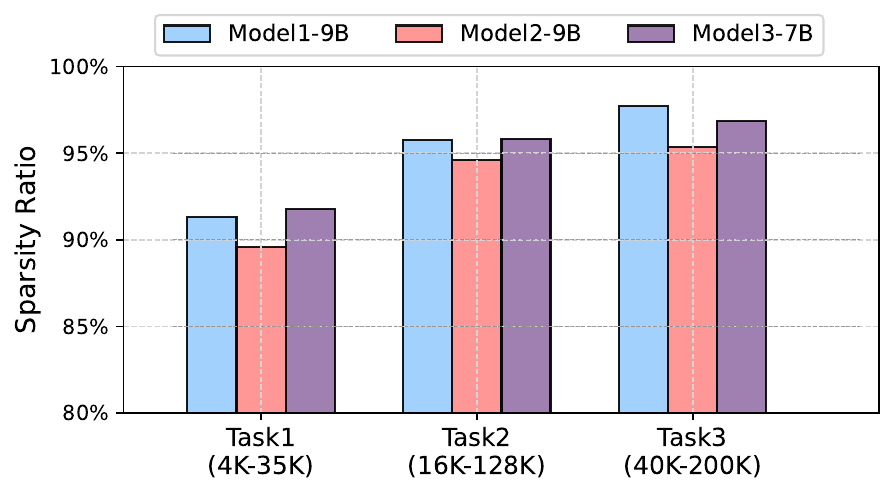}
  \vspace{-1.0em}
  \caption{The average sparsity ratio of three different models with long-context window on tasks with varying length ranges.} 
  \label{avg_sparsity}
  \vspace{-1.0em}
\end{figure}

\subsection{Inherently-High Attention Sparsity}

In attention computation, applying \texttt{softmax} over long sequences tends to reduce the influence of smaller elements, making them less significant. This insight motivates us to investigate the inherent sparsity in the attention scores, which can potentially accelerate the attention mechanism without compromising accuracy. Formally, the full attention score matrix $\textbf{P}$ can be approximated with sparse attention score $\hat{\textbf{P}}$,
\begin{equation}
\hat{\textbf{P}} = \texttt{softmax}(\frac{\textbf{QK}^{T}}{\sqrt{d}} - c(1-\textbf{M}))
\end{equation}
where $\textbf{M} \in \{0, 1\}^{S_q\times S_k}$ is a binary mask matrix that determines the sparse attention pattern, and $c$ is a large constant that effectively zeroes out masked attention scores after the \texttt{softmax} operation. The sparsity ratio measures the percentage of attention scores that are masked.

Our observations reveal that LLMs inherently exhibit a significant sparsity ratio in their attention computations, even without explicit optimization for this property. This finding emerges from our comprehensive evaluation of attention patterns across different model architectures and input prompts, as shown in Figure~\ref{avg_sparsity}. These observations suggest that attention sparsity is an intrinsic characteristic of how LLMs process information during the prefill phase, and the average sparsity ratio is high across different models and datasets, suppressing $89.6\%$.

\section{Motivation}
\label{sec:motiv}

While the high attention sparsity has been discussed in recent works~\cite{jiang2024minference,li2024snapkv,xiao2024duoattention}, we find that the adaptive sparsity ratio and dynamic sparse pattern make it challenging to effectively exploit this peoperty to accelerate prefill attention computation for long context LLMs.

\begin{figure}[t]
  \centering
  \includegraphics[width=0.85\linewidth]{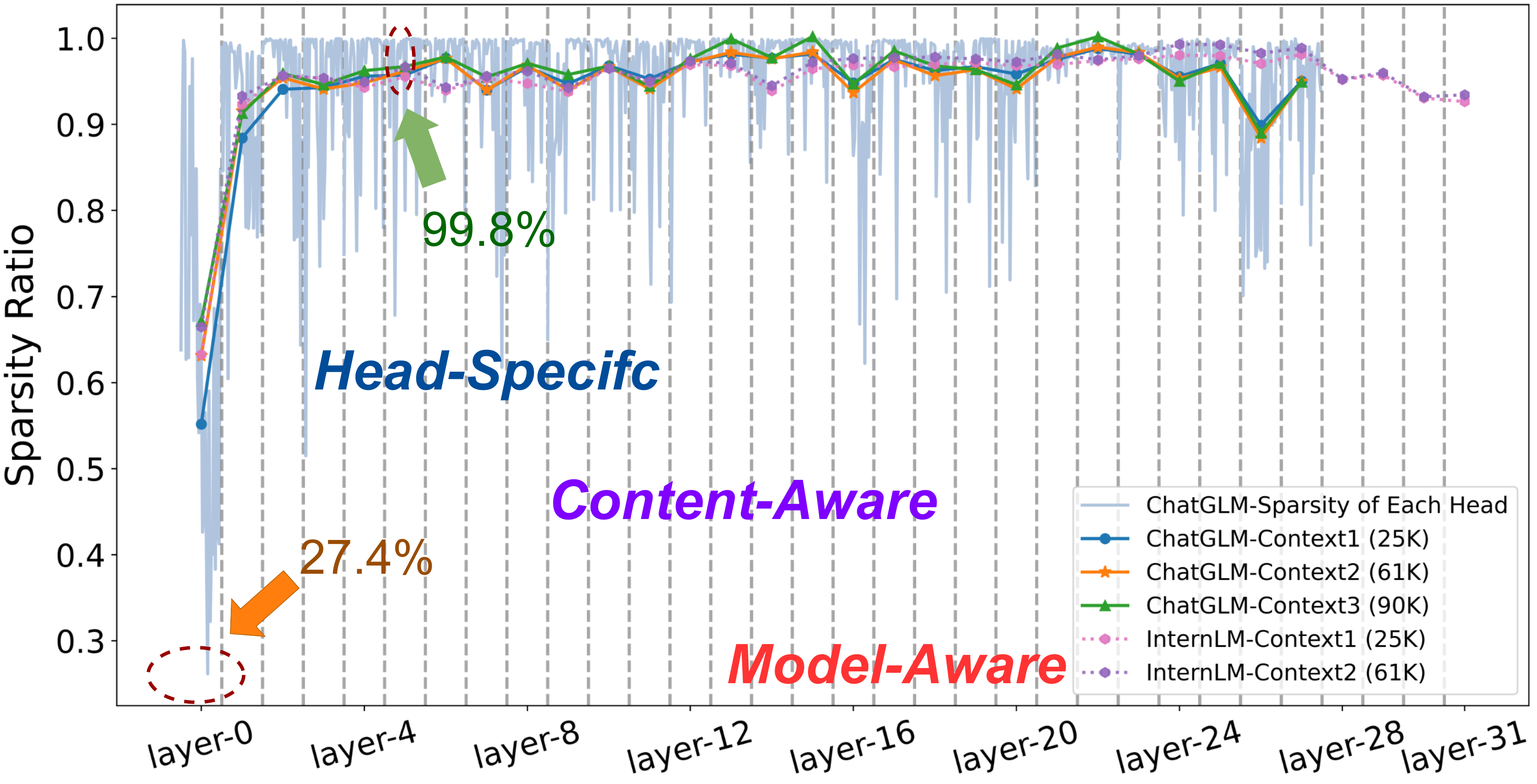}
  \caption{The sparsity ratio of ChatGLM3 (28 layers$\times$32 heads) and InternLM2 (32 layers$\times$32 heads), evaluated over different tasks during prefill. The sparsity ratio varies across different attention heads, input contents and model architectures. 
   }
  \label{obs}
\end{figure}

\begin{figure*}[t]
    \centering
   
    \begin{subfigure}[b]{0.25\textwidth}
        \centering
        \includegraphics[width=0.80\textwidth]{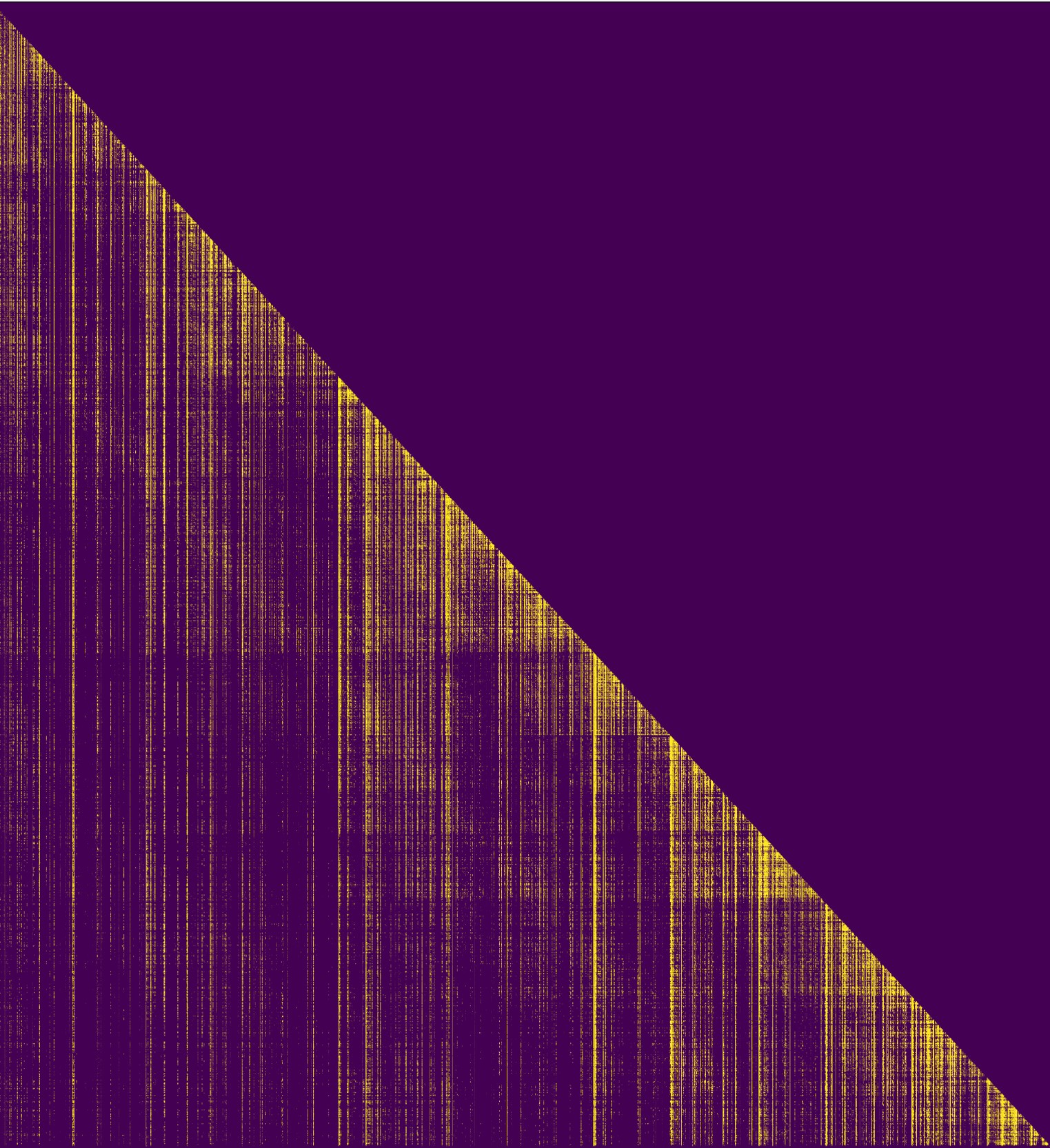}
        \caption{Head-A, prompt-A}
    \end{subfigure}%
    \begin{subfigure}[b]{0.25\textwidth}
        \centering
        \includegraphics[width=.80\textwidth]{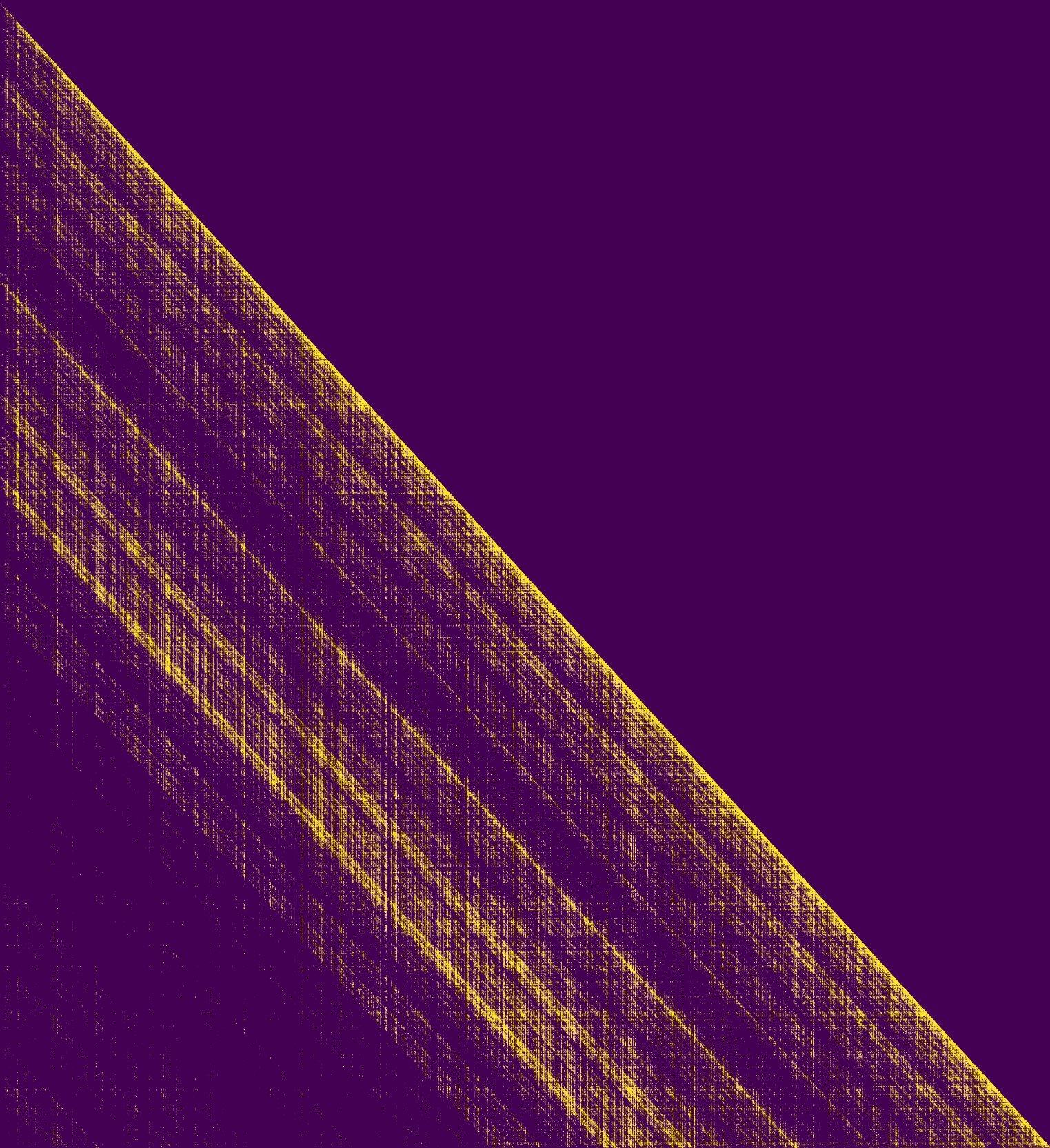}
        \caption{Head-B, prompt-A}
    \end{subfigure}%
    \begin{subfigure}[b]{0.25\textwidth}
        \centering
        \includegraphics[width=.8\textwidth]{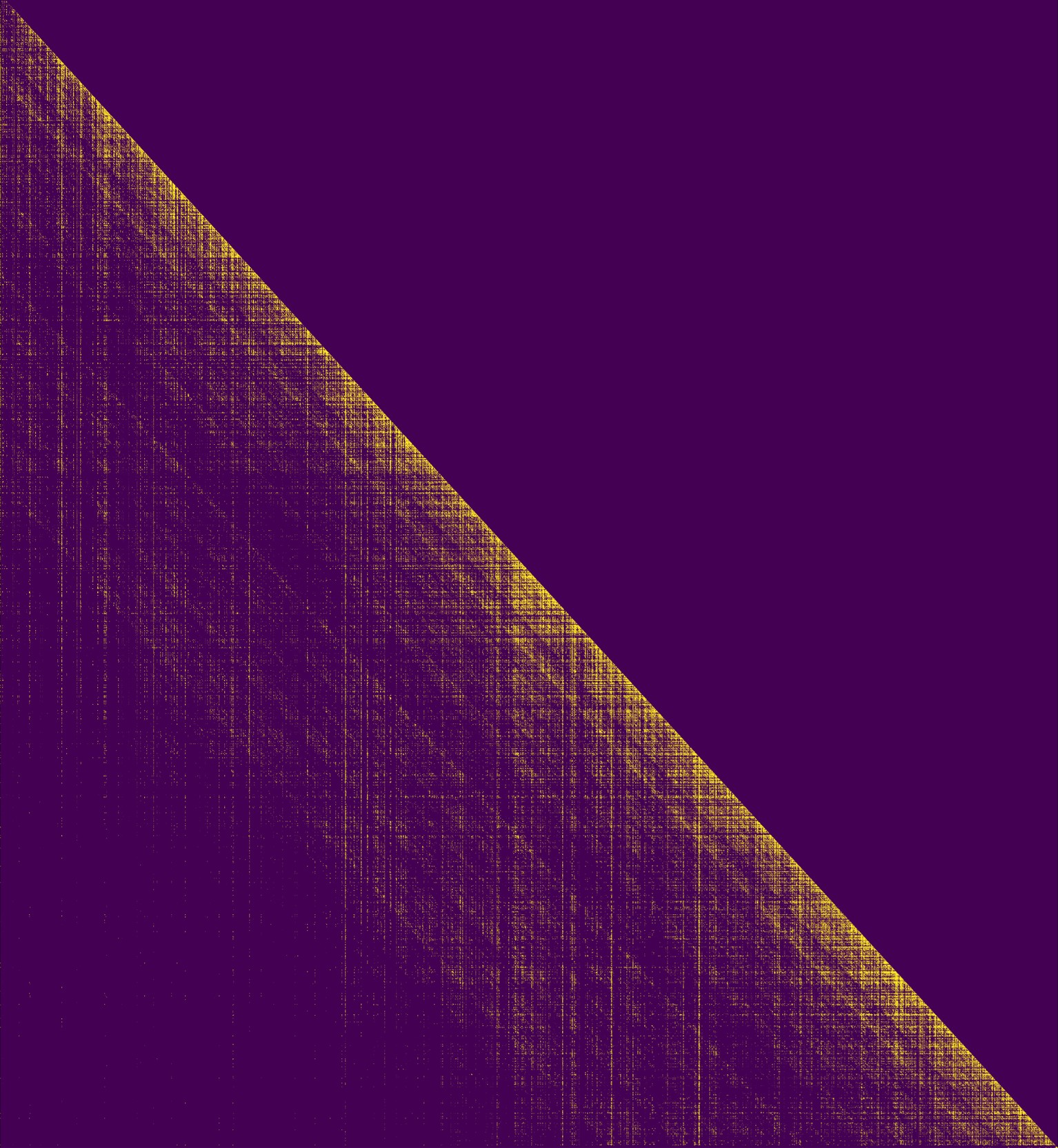}
        \caption{Head-C, prompt-A}
    \end{subfigure}%
    \begin{subfigure}[b]{0.25\textwidth}
        \centering
        \includegraphics[width=.8\textwidth]{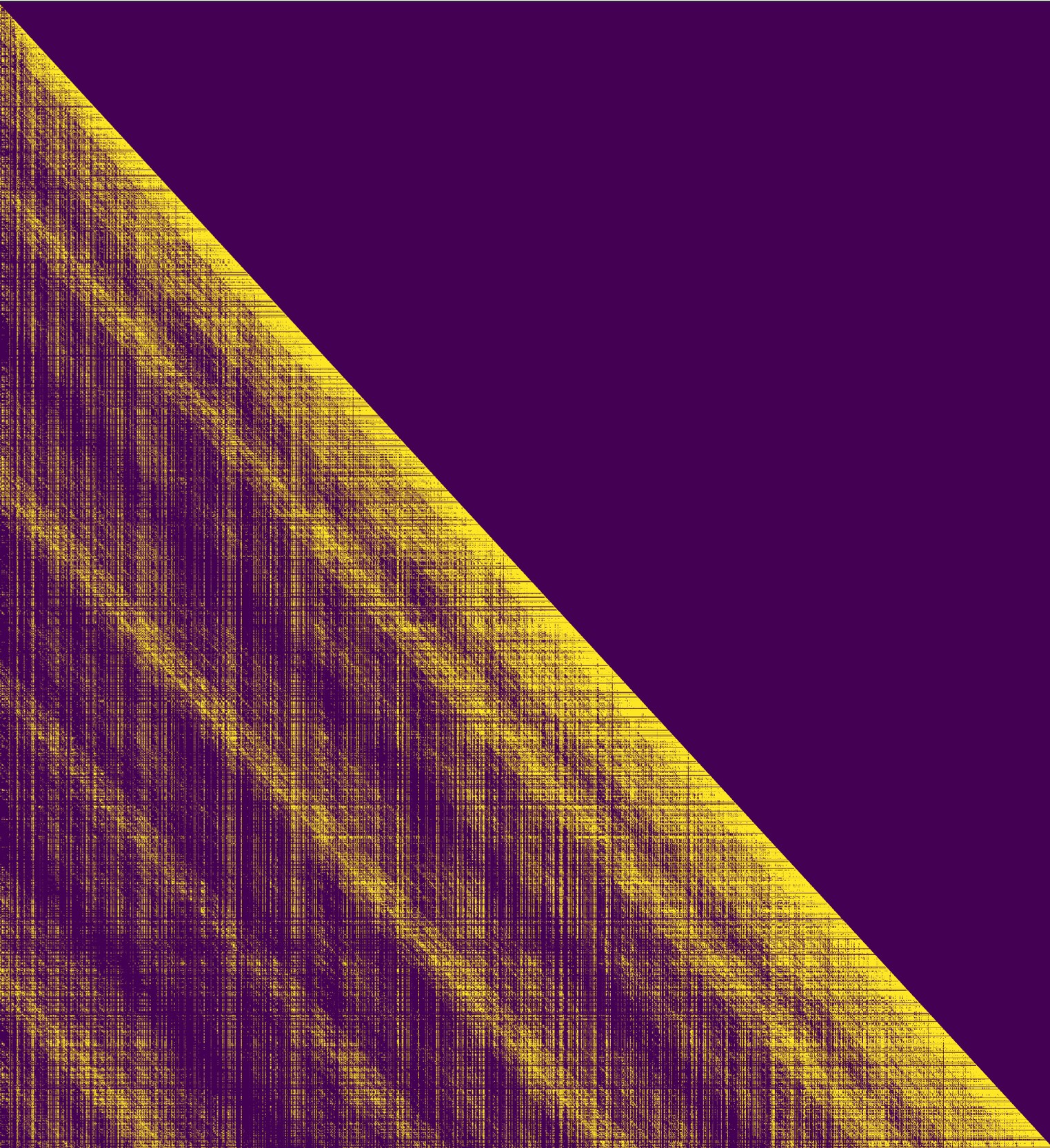}
        \caption{Head-C, prompt-B}
    \end{subfigure}
    \caption{The visualization of attention reveals diverse structured sparse patterns. Different heads with same prompt exhibit dynamic sparse indices and ratios, but the patterns can generally be categorized into (a) \textbf{column} , (b) \textbf{slash} , or  (c) \textbf{composed} pattern with column and slash. These structured patterns generally extend across the entire head. However, in some heads, such as (c), a prominent column structure in the upper half gradually fades in the lower half. Additionally, as shown in (d), the same head exhibits significant pattern differences under different prompts, highlighting the content-aware nature. }
    \vspace{-0.7em}
    \label{fig:sparse_pattern}
\end{figure*}

\subsection{C\#1: Adaptive Sparsity Ratio}

Our observations reveal that attention sparsity in LLMs exhibits adaptive sparsity ratio across three dimensions: attention heads, input contents, and model architectures (Figure~\ref{obs}).
\begin{itemize}
    \item \textit{Head-Specific:} different attention heads exhibit remarkably different sparsity ratios, even within the same layer. Speciafically, one head in the first layer has a sparsity ratio as low as $27.4\%$, while the highest can reach $99.8\%$.
    \item \textit{Content-Aware:} the sparsity ratios of different input prompts and context lengths are different. Our observation indicates that as the context becomes longer, the sparsity ratio increases correspondingly (Appendix~\ref{appendix-sparsity}).
    \item \textit{Model-Aware:} different models exhibit distinct sparsity ratios (Figure~\ref{avg_sparsity}, Figure~\ref{obs}, Appendix~\ref{appendix-sparsity}).
\end{itemize}

This adaptive sparsity ratio demonstrates that applying a fixed sparsification budget across all attention heads and input contents, as employed by MInference~\cite{jiang2024minference} and DuoAttention~\cite{xiao2024duoattention}, to search for sparse pattern is suboptimal. To maximize the efficiency of sparse attention while maintaining accuracy, the sparsity ratio need to be dynamically determined at runtime for each individual attention head and input prompt, allowing the model to adapt to the inherent variations in sparsity ratios.

\subsection{C\#2: Dynamic Sparse Pattern}
\label{dynamic-pattern}

Our observations also reveal that the attention pattern varies across different attention heads, input contents, and model architectures. Figure~\ref{fig:sparse_pattern} visualizes distinct sparse patterns from different heads. Our analysis identifies two significant sparse patterns that substantially contribute to the attention score. The \textbf{\textit{column pattern}} embodies crucial global contextual information (Figure~\ref{fig:sparse_pattern}(a)), with the attention sink~\cite{xiao2023efficient} as a typical example. On the other hand, the \textit{\textbf{slash pattern}} maintains connections between contexts of regular intervals (Figure~\ref{fig:sparse_pattern}(b)), such as the local window pattern which captures recent context information. These two patterns can be combined to cover diverse sparse patterns. For example, Figure~\ref{fig:sparse_pattern}(c-d) display a composition of column and slash patterns on a single head. Moreover, the concrete sparse pattern varies across different input contents.

These dynamic patterns present a significant challenge to determine the exact sparse pattern during inference. While MInference~\cite{jiang2024minference} acknowledges similar column and slash patterns, its approach of classifying attention heads into fixed categories (A-shape, Vertical-Slash, or Block-Sparse) and determining optimal patterns offline fails to capture the content-dependent variations in sparse patterns. This limitation highlights the need for a more flexible, runtime-adaptive approach to pattern selection.

\begin{figure}[t]
    \centering
   
    \begin{subfigure}[b]{0.25\textwidth}
        \centering
        \includegraphics[width=0.94\textwidth]{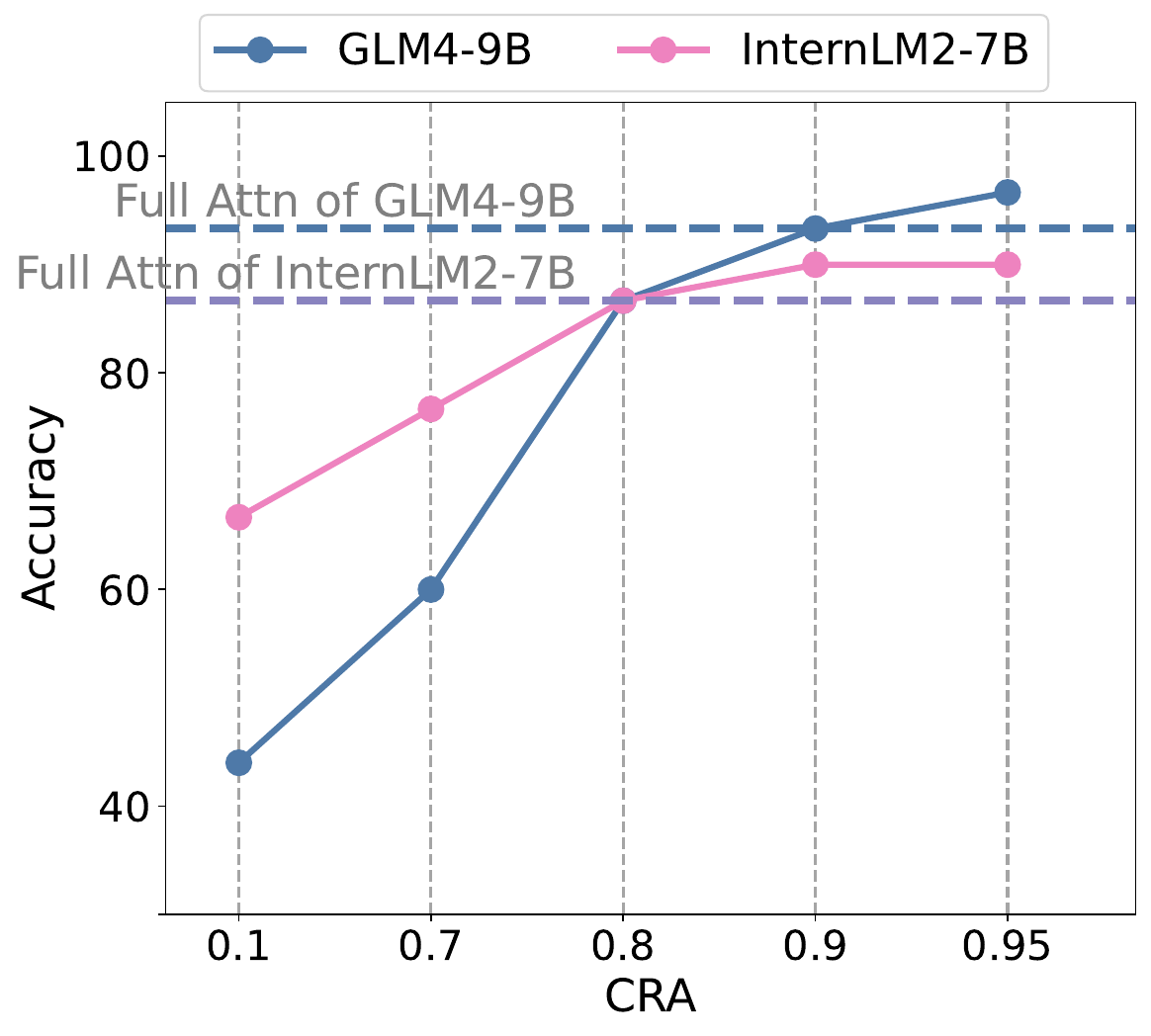}
        \caption{Task1: Retrieval }
    \end{subfigure}%
    \begin{subfigure}[b]{0.25\textwidth}
        \centering
        \includegraphics[width=.94\textwidth]{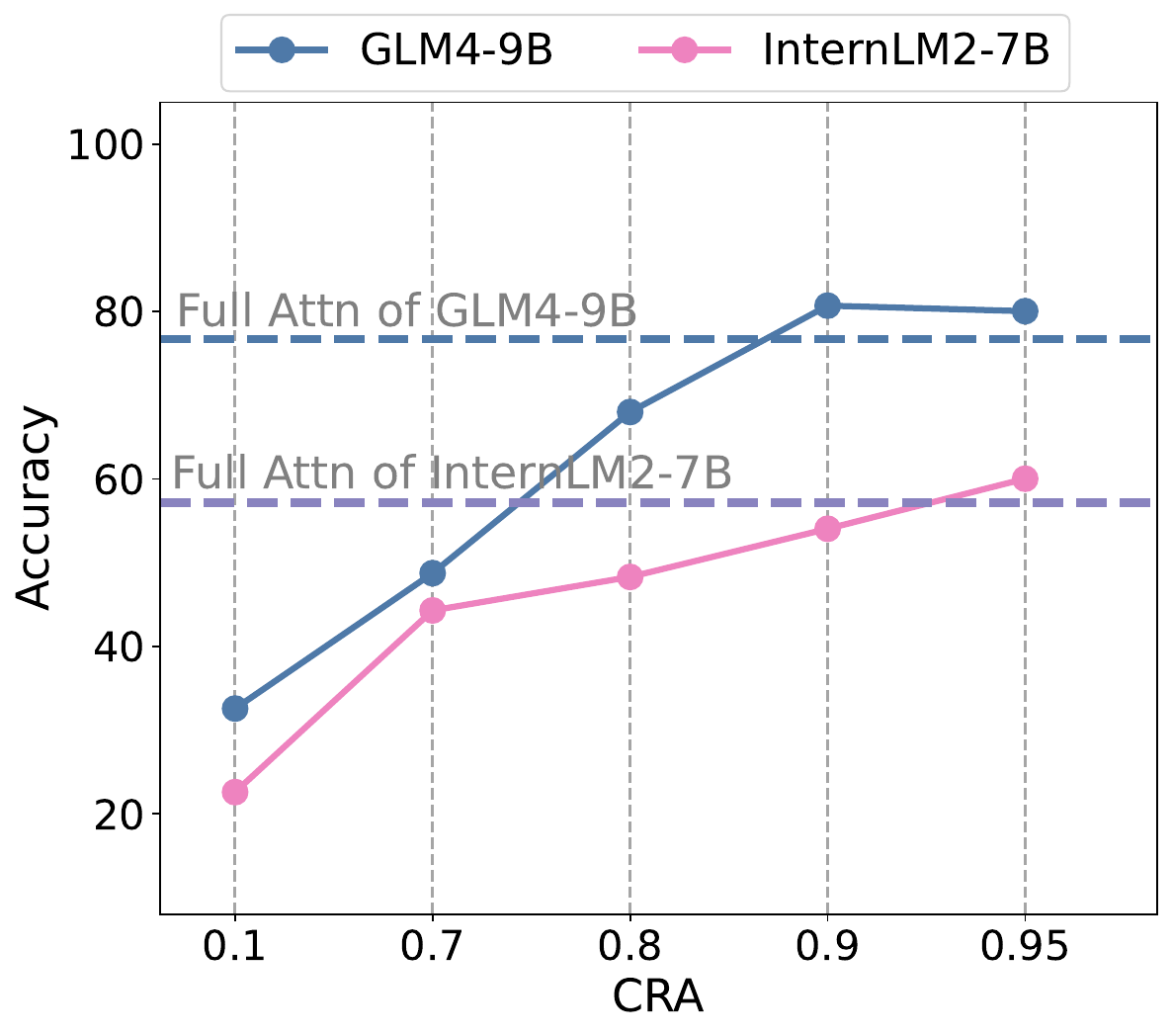}
        \caption{Task2: Question Answering}
    \end{subfigure}%
    \caption{The curves of model accuracy and attention recall with changing \textbf{CRA} thresholds $\alpha$ for different tasks. }
    \vspace{-1.3em}
    \label{fig:cra_accuracy}
\end{figure}

\begin{figure*}[t]
  \centering
  \includegraphics[width=1.00\linewidth]{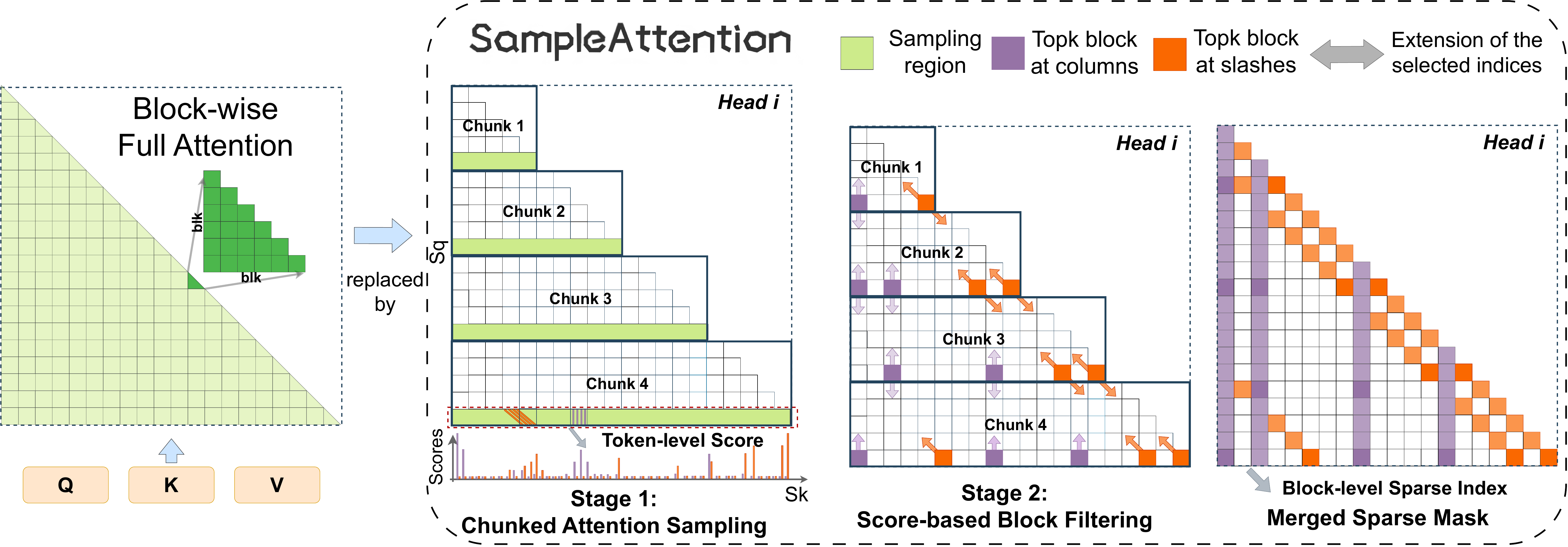}
  \caption{\WORK{} replaces the original full attention with a two-stage implementation. In the first stage, token-level attention scores are computed by performing chunked sampling across multiple query blocks and accumulating the scores along the column and slash direction. In the second stage, we determine the minimum quota of blocks required for each sampling region based on the block-reduced scores and  thresholds \((\alpha_{c},\alpha_{s})\). Then, we perform \texttt{top-k} operation on each head to filter out the required block indices \(I_{c}\) and \(I_{s}\). These indices will be extended along the column and slash patterns and merged into \(\hat{\textbf{M}}\) to enable sparse computation in attention.}
  \label{fig:overview}
  \vspace{-1em}
\end{figure*} 

\subsection{Insight: Cumulative Residual Attention}

To effectively balance the trade-off between efficiency and accuracy when determining runtime sparsity ratios and patterns, we need a reliable metric to guide these decisions. Our finding reveals that the \textit{\textbf{Cumulative Residual Attention}} (\textbf{CRA}), defined as the minimum sum of remaining attention probabilities per query after sparsification, serves as a robust indicator of model accuracy. As demonstrated in Figure~\ref{fig:cra_accuracy}, there exists a consistent positive correlation between the \textbf{CRA} threshold and model accuracy across different LLMs and tasks. This relationship provides a principled way to navigate the efficiency-accuracy trade-off: while lower \textbf{CRA} thresholds enable greater computational speedups, they should be carefully balanced against potential accuracy degradation. 

Leveraging this insight, we can dynamically identify a minimal set of attention indices that satisfy the desired \textbf{CRA} threshold, thereby optimizing computational efficiency while preserving accuracy. This dynamic selection approach naturally accommodates varying sparsity ratios and sparse patterns across different attention heads, input contents, and model architectures. 

\begin{figure}[t]
  \centering
  \includegraphics[width=0.85\linewidth]{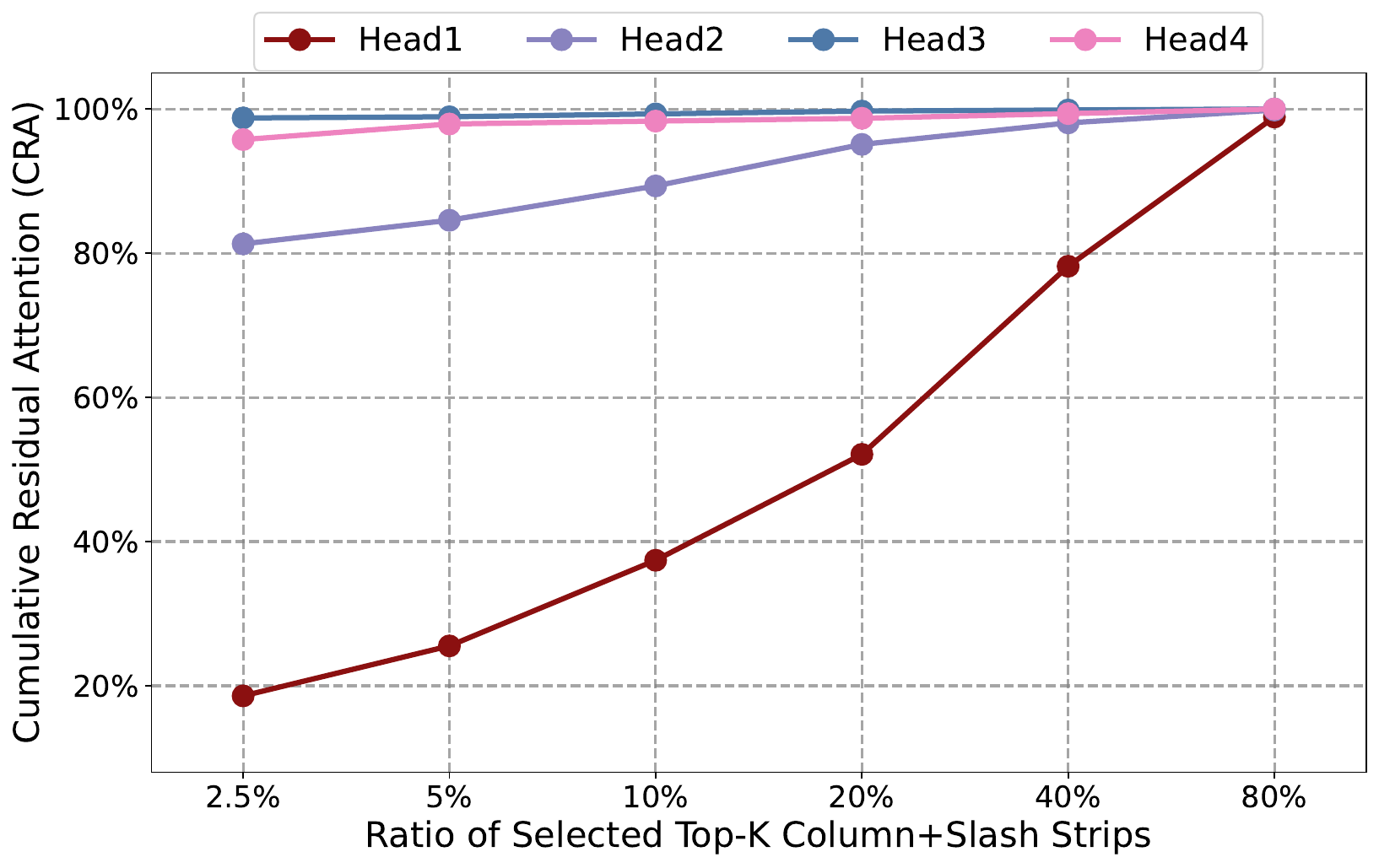}
  \caption{Relationship between the ratio of selected \texttt{top-k} columns-slashes and \textbf{CRA}. The high similarity of the numerical distribution  enables a small amount of critical stripes-slashes to cover the majority values of the full attention score matrix.
   } 
  \label{fig:cra_ratio}
\end{figure}

The precise dynamic selection for a \textbf{CRA} threshold requires computing the full attention score, which is computationally expensive. Inspired by the observed significant column and slash pattern, we find that selecting an appropriate number of key column and slash strips can accurately approximate the \textbf{CRA} (Figure~\ref{fig:cra_ratio}). The number of selected strips can be determined at runtime to vary the sparsity ratio across different heads and contents. Additionally, the combination of column and slash patterns provides sufficient flexibility to capture diverse attention distributions encountered.

\section{\WORK{}}
\label{sec:method}
In this section, we demonstrate how \WORK{} efficiently determines dynamic sparsity ratios and structured patterns at runtime, ensuring maximum efficiency through sparse acceleration while maintaining nearly lossless accuracy. 
In Section~\ref{sec-overview}, we provide an overview of the key considerations and the two-stage implementation of \WORK{}. 
Subsequently, in Section~\ref{stage1} and Section~\ref{stage2}, we delve into the details of the two critical stages of \WORK{}: query-guided chunked sampling and score-based key-value filtering.
Additionally, in Section~\ref{hyper-tune}, we explain how the introduced hyperparameters affect the tradeoff between accuracy and performance, and outline the method for tuning them.
Finally, we detail the hardware-efficient implementation in Section~\ref{sec-hardware-efficient}.

\subsection{Overview}
\label{sec-overview}

Given a desired \textbf{CRA} threshold, \WORK{} dynamically selects a set of important column and slash strips to capture adaptive sparsity ratio and dynamic sparse pattern. However, the selection is non-trivial since \WORK{} requires fast and accurate estimation of attention scores. To address this, \WORK{} proposes a two-stage sampling algorithm.

The first stage is \textit{Query-Guided Chunked Sampling}, where \WORK{} estimates the full attention score by computing the attention scores for a few queries, inspired by the observed column pattern. While prior work like MInference~\cite{jiang2024minference} uses only last $last_q$ queries for estimation, we found this approach insufficient for capturing complex hybrid column and slash patterns. For instance, as shown in Figure~\ref{fig:sparse_pattern}(c), the column pattern is significant in the upper half region, but gradually fades in the lower half region. To capture such varying patterns effectively, \WORK{} partitions the queries into $chunk_n$ equal segments and perform separate attention sampling for each chunk, enabling more accurate pattern detection across the entire attention matrix.

The second stage is \textit{Score-Based Key-Value Filtering}, where \WORK{} filters key columns and slashes that meet a specified \textbf{CRA} threshold $\alpha$ based on the sampled attention scores. A naive approach of jointly selecting column and slash strips would require evaluating $n_{column}\times n_{slash}$ combinations, making the selection process computationally expensive. To improve efficiency, we decompose the single threshold $\alpha$ into separate thresholds: $\alpha_c$ for columns and $\alpha_s$ for slashes. This decomposition reduces the computational complexity from $n_{column}\times n_{slash}$ to $n_{column} + n_{slash}$. The filtering stage independently selects column and slash strips before merging them into the final pattern.

\WORK{} introduces several tunable hyperparameters to control the trade-off between efficiency and accuracy. To facilitate rapid adoption of \WORK{}, we provide a method for automatic hyperparameter tuning on a compact validation dataset, which maximizes computational efficiency while maintaining accuracy.

\begin{algorithm}[H]
\captionsetup[algorithm]{singlelinecheck=off}
\caption{Two-stage Implementation  of \WORK{}}
\label{alg:code}
\begin{algorithmic}
  \STATE {\bfseries Input:} $\boldsymbol{Q}   ,\boldsymbol{K},\boldsymbol{V} ,\boldsymbol{\alpha_{c}},\boldsymbol{\alpha_{s}} \in [0,1],\boldsymbol{chunk_{n}}$

    \LineComment{Stage1: Query-Guided Chunked Attention Sampling}
    \STATE $\boldsymbol{itv} \gets \frac{\boldsymbol{S_q}}{\boldsymbol{chunk_{n}}}, \; \boldsymbol{blk} \gets 128$\\
    \STATE $\boldsymbol{Q_{slice}} \gets [\boldsymbol{Q}_{\mathrm{[i*itv-blk:i*itv]}} \; \mathrm{for} \; \boldsymbol{i}  \; \mathrm{in} \; \mathrm{range} ( 1, \boldsymbol{chunk_{n}}+1)] $\\
    \STATE $\boldsymbol{\hat{A}} \gets \mathrm{softmax}\left(\boldsymbol{Q}_{slice} \bm{K}^{\top} / \sqrt{d} + \bm{m}_{\text{casual}} \right)$
    % \STATE $\boldsymbol{\hat{A}} \gets \mathrm{softmax}\left(\boldsymbol{Q}_{[\boldsymbol{itv}-\boldsymbol{block_{k}}::\boldsymbol{itv}]} \bm{K}^{\top} / \sqrt{d} + \bm{m}_{\text{casual}} \right)$

    \STATE $\boldsymbol{\hat{A}_{c}}, \boldsymbol{\hat{A}_{s}} \gets \mathrm{block\_reduction}\left(\boldsymbol{\hat{A}},  \boldsymbol{blk} \right)$
    
    % \LineComment{Indices of top $k_v$ vertical line, sum in vertical}
    \LineComment{Stage2: Score-Based Key-Value Block Filtering}
    \STATE $\boldsymbol{k}_c \gets \mathrm{find\_k}\left(\mathrm{cumsum} (\mathrm{sort}(\boldsymbol{\hat{A_{c}}})), \boldsymbol{\alpha_c} \right)$
    \STATE $\boldsymbol{I}_c \gets \mathrm{arg\_topk}\left( \boldsymbol{\hat{A_c}} , \boldsymbol{k_c} \right) $
    
    \STATE $\boldsymbol{k}_s \gets \mathrm{find\_k}\left(\mathrm{cumsum} (\mathrm{sort}(\boldsymbol{\hat{A_{s}}})), \boldsymbol{\alpha_s} \right)$
    \STATE $\boldsymbol{I}_s \gets \mathrm{arg\_topk}\left( \boldsymbol{\hat{A_s}} , \boldsymbol{k_s} \right) $

    \LineComment{Extend and Merge Block-sparse Mask across Each Head}
    \STATE $\boldsymbol{\hat{M}} \gets \mathrm{merge\_index}(\boldsymbol{I}_c, \boldsymbol{I}_s,  \boldsymbol{itv} )$

    \LineComment{Final Sparse FlashAttention with Block Index}
    \STATE $\boldsymbol{O} \gets \mathrm{sparse\_flash\_attn}\left( \boldsymbol{Q},  \boldsymbol{K},  \boldsymbol{V} , \boldsymbol{\hat{M}}\right)$

    \STATE $\mathrm{return}\,\,\,\boldsymbol{O}$
   
\end{algorithmic}
\end{algorithm}

\subsection{Stage1: Query-Guided Chunked Sampling}
\label{stage1}
As discussed in Section~\ref{sec-overview}, we need to divide the queries into chunks to ensure that sampling can more accurately determine the sparse structure within each head.
Compared to random sampling or bottom sampling, this equidistant sampling technique is low-overhead and more stable.
Our experiments show that this straightforward approach is effective: sampling a small number of query blocks can accurately approximate the actual \textbf{CRA} (Further details can be found in Appendix~\ref{appendix-sampling}).
It should be noted that existing methods~\cite{jiang2024minference,li2024snapkv} commonly use bottom sampling, which fixes \(chunk_{n}\) at 1. This means that only query blocks at the bottom of the score matrix are sampled. Due to the overly concentrated sample locations, this approach may lead to biases in index selection. 

Subsequently, we accurately compute the attention scores for the query blocks at the bottom of each chunk. These token-level scores are then reduced at the block size ($blk$) granularity in both column and slash directions. 
The resulting block-level scores help us dynamically determine the required sparsity ratios and indices in the second stage.

\subsection{Stage 2: Score-Based Key-Value Filtering}
\label{stage2}
In the second stage, our approach performs a refined filtering of key-value indices based on the sample attention scores. 
A significant challenge remains in efficiently and dynamically selecting the minimal set of key-value indices of interest under different patterns, denoted as \( I_{c} \) and  \( I_{s} \) , that align with the input prompt and satisfy the CRA threshold \(\alpha\). 
Determining a fixed quota of critical indices does not achieve the optimal balance between accuracy and performance, especially given the extensive sequence lengths. To address this inefficiency, we employ an adaptive approach that filters block-level indices based on blocked attention scores in both the column and slash directions, comparing them with their respective thresholds.

In detail, as shown in Algorithm~\ref{alg:code}, for each sampled block within a chunk, \WORK{} first accumulates the attention scores along the column and slash directions and then reduces them at the block granularity. This accumulation serves as a statistical approximation of the overall attention scores.
Using these block-level scores in both directions, \WORK{} can effectively select the minimal number of key-value blocks \(k_c\) and \(k_s\) that meet the CRA thresholds \(\alpha_c\) and \(\alpha_s\) for each attention head, respectively.
Finally, based on the derived minimal quota, \WORK{} performs a \texttt{top-k} operation in each direction to filter out the essential block indices \(I_c\) and \(I_s\).
It is noteworthy that sampled chunks from different positions may filter out different sets of column and slash indices. \WORK{} extends these indices obtained from each sampled block according to the patterns so that they cover the entire attention matrix. This ensures that the final merged mask \(\hat{M}\) achieves an almost lossless sparse approximation.
This approach also enables the identification of critical attention sinks and local window masks, thereby maintaining stable accuracy.

\subsection{Hyperparameter Tuning} 
Due to the introduction of three hyperparameters in \WORK{}, as shown in Table~\ref{tab:hyperparameter}, it is crucial to analyze their impact on the accuracy and performance. Here, we briefly discuss the influence of these parameters and outline the approach to tuning them. Detailed results of modifying these hyperparameters are studied in Section~\ref{ablation-study}. 

\textit{Thresholds for Columns and Slashes.} The most critical hyperparameters in SampleAttention are the CRA thresholds \(\alpha_c\) and \(\alpha_s\). Generally, larger threshold values can reduce the speedup but enhance the model's accuracy. 
Therefore, it is necessary to predetermine cost-effective threshold values for columns and slashes offline using a compact dataset. We can use the accuracy and latency metrics of FlashAttention2~\cite{dao2023flashattention} as reference standards for this process. Moreover, since long-context tasks span a wide range, further segmenting the context by length and tuning each segment individually can more effectively leverage sparsity at different lengths, thereby achieving more cost-effective threshold values.
 
\textit{Sampling Positions and Ratios.} 
Additionally, the number of sampling chunks significantly affects \WORK{}'s performance by influencing the sampling positions and the ratio of selected indices.
For instance, too few sampling samples may fail to capture the full sparse structures in the head, thereby reducing accuracy. Conversely, excessive sampling can increase overhead and introduce redundant computations in attention. Therefore, during the tuning, we introduce multiple values for \(chunk_{n}\) to expand the search space and identify more efficient configurations.

\label{hyper-tune}
\begin{table}[h!]
  \centering
  \small
  \caption{The meaning of hyperparameters and they will be tuned offline for different length ranges.}
  \label{tab:hyperparameter}
  \begin{tabular}{c|c}
   \toprule
    Hyperparameter & Description  \\
    \midrule
    \textbf{$\alpha_{c}$} & The desired \textbf{CRA} threshold for columns   \\
     \textbf{$\alpha_{s}$} & The desired \textbf{CRA} threshold for slashes  \\
     \textbf{$chunk_{n}$} & The number of sampling chunks  \\
    \bottomrule
  \end{tabular}
\end{table}

\subsection{Hardware-efficient Implementation}
\label{sec-hardware-efficient}
To achieve substantial speedup in wall-clock time, \WORK{} is implemented with IO-awareness to maximize hardware-efficiency. First, the query-guided key-value filtering involves a series of small operators (\texttt{bmm}, \texttt{mask\_fill}, \texttt{softmax},  \texttt{reduction}) that read and write large intermediate results. \WORK{} significantly reduces IO overhead by fusing these operators. Second, \WORK{} implements an efficient adaptive structured sparse attention kernel by modifying FlashAttention2~\cite{dao2023flashattention}. These hardware-aware optimizations enhance speed performance significantly.

\section{Experiments}
\label{experiments}

\begin{figure*}[htp]
  \centering
  \includegraphics[width=1.00\linewidth]{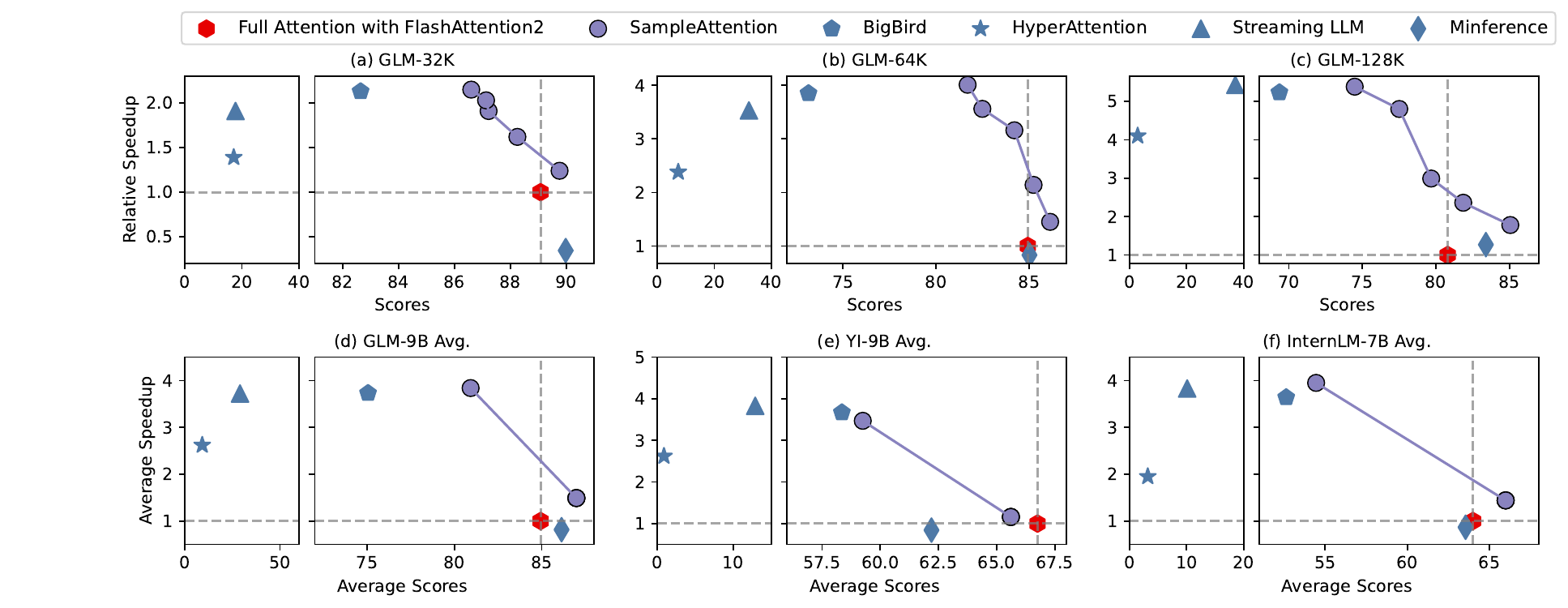}
  \caption{The trade-off between accuracy (measured via RULER benchmark scores) and speedup (relative to FlashAttention2) is analyzed across various sparse attention methods. Figures (a)–(c) illustrate this trade-off under the GLM4-9B architecture for sequence lengths of 32K, 64K, and 128K, respectively. Meanwhile, Figures (d)–(f) present averaged results across different models for each of the three sequence lengths, focusing exclusively on configurations demonstrating optimal performance in both acceleration and precision. Notably, our proposed method consistently outperforms MInference in both accuracy and speedup across all evaluated scenarios, showcasing superior efficiency-accuracy trade-offs.}
  \label{tradeoff-result}
\end{figure*}

\subsection{Setup}

\textbf{Backbones.}
We evaluate our method on three widely used open-source LLM variants: \textbf{ChatGLM4-9B} with a 1M context window based on GLM~\cite{du2021glm,glm2024chatglm}; ; \textbf{YI-9B}, featuring a 200K context window~\cite{young2024yi}; and \textbf{InternLM2-7B}, also with a 200K context window~\cite{cai2024internlm2}.
All utilized models are decoder-only transformers~\cite{radford2018improving}, and are pre-trained via causal language modeling.  They encompass similar architectural components, such as rotary positional encoding~\cite{su2024roformer}, and grouped-query attention~\cite{ainslie2023gqa}. Simultaneously, there are notable differences, e.g., the former augments the context window capacity via continued training with an extended sequence length, whereas the latter achieves length extrapolation through rope scaling.
It is important to note that we specifically replace the full-attention implementation during the prompt prefill stage with \WORK{} and baseline methods, while preserving an uncompressed key-value (KV) cache and dense attention computation in the decoding phase.

\textbf{Tasks.} 
We evaluate \WORK{} and other methods' understanding capabilities in long-context scenarios on three distinct tasks: RULER~\cite{hsieh2024ruler}, LongBench~\cite{bai2023longbench} and Infinite Bench~\cite{zhang2024bench}. 
\textbf{RULER} provides a comprehensive evaluation of long-context language models through flexible configurations  for sequence lengths. Unlike the traditional needle-in-a-haystack~\cite{kamradt2023needle} test,  RULER extends it by incorporating diverse types and quantities of "needles" and introduces new tasks such as multi-hop tracing and aggregation, which evaluate more complex behaviors beyond simple retrieval. RULER encompasses 13 tasks, making it an excellent tool for testing long-context understanding.
\textbf{LongBench}, a multi-task benchmark, comprises single and multi-document QA, summarization, few-shot learning, synthetic tasks, and code completion. It offers over 4,750 test cases with task lengths from  4K-35K.
\textbf{InfiniteBench}, a benchmark specifically designed to evaluate language models' ability in handling, understanding, and reasoning in contexts exceeding an average length of 200K. It comprises 10 unique tasks, each crafted to assess different aspects of language processing and comprehension in extended contexts.

\textbf{Baselines and settings.} 
We conducted all experiments on a single NVIDIA-A100 GPU (80GB) to evaluate accuracy and performance of attention operation during the prefill stage. 
We consider the full attention (as the gold baseline), Minference~\cite{jiang2024minference}, BigBird~\cite{zaheer2020big}, Streaming-LLM~\cite{xiao2023efficient}, HyperAttention~\cite{han2023hyperattention} and Hash-Sparse~\cite{pagliardini2023faster} as baselines to compare model accuracy across different tasks. 
Minference requires pre-profiling to determine the optimal pattern.
And both BigBird and StreamingLLM were assigned a window size ratio of $8\%$. 
BigBird retains a global ratio of $8\%$. 
StreamingLLM sets its initail attention sink at 4 tokens.
HyperAttention set both bucket size and the number of sampled columns to 256.
For \WORK{}, we generate small-scale tasks using the RULER benchmark at specific lengths (e.g., 16K, 32K, 64K, 128K) to tune hyperparameters and subsequently apply these parameters to tasks across different length ranges. This approach enables \WORK{} to achieve an optimal trade-off balance within each length range.

\subsection{Trade-off between Accuracy and Efficiency}
\label{perf_result}

\begin{table*}[ht]
  \caption{Accuracy comparison across various sparse methods on LongBench and InfiniteBench under GLM4-9B model. The hyperparameters applied are the optimal accuracy configurations, tuned for different sequence length ranges. The best results are highlighted in \textbf{Bold} while the second best results are marked with an \underline{Underline}.}
  \label{tasks-long}
  \begin{adjustbox}{max width=\textwidth, left}
    \begin{tabular}{c|c|c|c|c|c|c|c|c|c|c|c}
      \toprule
      \multirow{3}{*}{Benchmark} & \multirow{3}{*}{Baseline} & \multicolumn{10}{c}{Task Type}   \\ 
      \cmidrule(lr){3-12}
       & & \makecell[c]{Single-\\Doc QA} & \makecell[c]{Multi-\\Doc QA} & \makecell[c]{Summari-\\zation} & \makecell[c]{Few-shot\\ Learning} & \makecell[c]{Synthetic\\ Tasks} & \makecell[c]{Code\\ Completion} & \multicolumn{4}{c}{Total Score} \\
      \midrule

      \multirow{6}{*}{\makecell[c]{LongBench}} & Full Attention & \underline{213.12} & \underline{174.35} & \underline{109.69} & 273.87 & 231.49 & 121.52 & \multicolumn{4}{c}{1124.04} \\
      & Ours & \textbf{214.53} & \textbf{174.42} & 108.92 & \textbf{278.33} & \textbf{234.55} & \textbf{125.18} & \multicolumn{4}{c}{\textbf{1135.93}} \\
      & Minference & 212.14 & 173.37 & \textbf{110.02} & \underline{274.45} & \underline{231.87} & \underline{124.37} & \multicolumn{4}{c}{\underline{1126.22}} \\
        & BigBrid & 207.57 & 146.45 & 95.64 & 272.17 & 161.60 & 117.38 & \multicolumn{4}{c}{1000.81}\\
        & StreamingLLM & 142.79 & 129.36 & 89.71 & 168.13 & 19.70 & 98.43 & \multicolumn{4}{c}{648.12} \\
        & HyperAttention & 125.74 & 119.08 & 88.05 & 206.35 & 32.69 & 86.35 & \multicolumn{4}{c}{658.26}\\
  \toprule
       & & \makecell[c]{En.Sum} & \makecell[c]{En.QA} & \makecell[c]{En.MC} & \makecell[c]{En.Dia} & \makecell[c]{Zh.QA} & \makecell[c]{Code.\\Debug} & \makecell[c]{Math.\\Find} & \makecell[c]{Retr.\\PassKey} & \makecell[c]{Retr.\\Number} & \makecell[c]{Retr.\\KV} \\

      \midrule
       \multirow{3}{*}{\makecell[c]{InfiniteBench}} & Full Attention  & \textbf{28.30} & \underline{12.17} & 58.95 & \textbf{34.00} & 13.22 & 30.71 & \underline{37.71} & \textbf{100} & \textbf{100} & \underline{44.0} \\

        & Ours  &\textbf{28.30} & \textbf{16.52}& \textbf{61.57} & \underline{31.50} & \underline{14.28} & \underline{31.40} & 37.14 & \textbf{100} & \textbf{100} & \textbf{49.6} \\

        & Minference  &28.00 & 11.39 &\underline{60.26} & 28.70 & \textbf{14.81} & \textbf{31.70} & \textbf{39.43} & \textbf{100} & \textbf{100} & 43.0 \\
      \bottomrule
    \end{tabular}
  \end{adjustbox}
\end{table*}

\textbf{Main results.} 
Figure~\ref{tradeoff-result} compares the trade-offs between accuracy and the speedup relative to FlashAttention2 among different sparse methods and \WORK{} under various hyperparametres on the RULER benchmark. Table~\ref{tasks-long}, on the other hand, compares the accuracy of different sparse methods on LongBench and InfiniteBench tasks.
The results show that:

\begin{itemize}
    \item The performance in accuracy of \WORK{} is consistently robust across all benchmarks (including subdomains), various models, and diverse sequence lengths. When compared to full attention, which serves as the gold standard, \WORK{} consistently achieves scores above 99\% of full attention, demonstrating near-lossless efficiency. Furthermore, our approach establishes a new Pareto frontier in the accuracy-efficiency trade-off, surpassing existing methods in both dimensions.
    \item While Minference achieves accuracy comparable to FullAttention at lengths of 32K and 64K, it fails to provide any speedup benefits. In contrast, sample attention maintains nearly lossless performance and achieves speedup ranging from $1.24\times$ to $2.36\times$ compared to FlashAttention2 across different lengths.
    \item BigBird exhibits varying degrees of performance degradation across different models and lengths. Nonetheless, on average, BigBird still achieves approximately 86\% of the scores achieved by full attention and provides a relatively stable speedup due to the nature of its static pattern.
    \item StreamingLLM and HyperAttention  result in performance degradation across all tasks, demonstrating that these techniques fail to capture critical KV elements in long sequences at the prefill stage. 
\end{itemize}

\subsection{Ablation Study and Tuning for Hyperparameter}
\label{ablation-study}

\begin{figure*}[htp]
  \centering
  \includegraphics[width=1.00\linewidth]{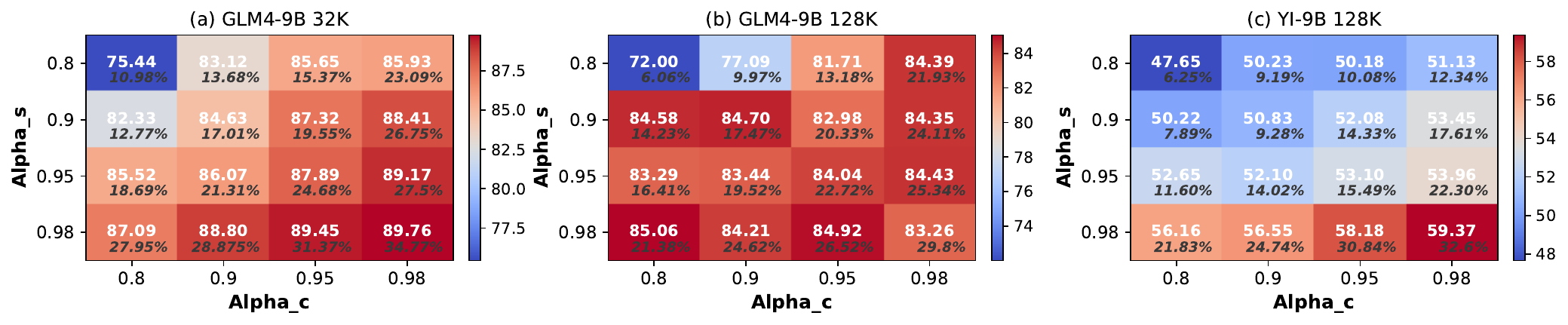}
  \caption{The heatmaps under different cases illustrate the impact of choosing different values of \(\alpha_{c}\) and \(\alpha_{s}\) on the accuracy of \WORK{} (the white text represents the scores under the RULER benchmark, while the black italic text denotes the actual ratio of calculated blocks). The \(chunk_{n}\) values for the GLM and YI models are set to 2 and 4, respectively.} 
  \label{threshold-result}
\end{figure*}

As introduced in Section~\ref{hyper-tune}, the sparse properties of different models exhibit significant variability across sequence lengths.
Therefore, we implement an automated, offline hyperparameter tuning method for the specified models on the small-scale datasets. This approach discretizes sequence lengths into distinct intervals and performs multi-task tuning within each segment. Such a strategy ensures an optimal balance between accuracy and efficiency over the entire range of sequence lengths.
Figure~\ref{threshold-result} demonstrates how varying the CRA thresholds for column and slash patterns influences both accuracy and the sparsity ratio under different conditions. Table~\ref{chunk-study}, on the other hand, investigates how different numbers of sampling chunks impact accuracy and computational speedup.

\begin{figure}[h]
  \centering
  \includegraphics[width=1.00\linewidth]{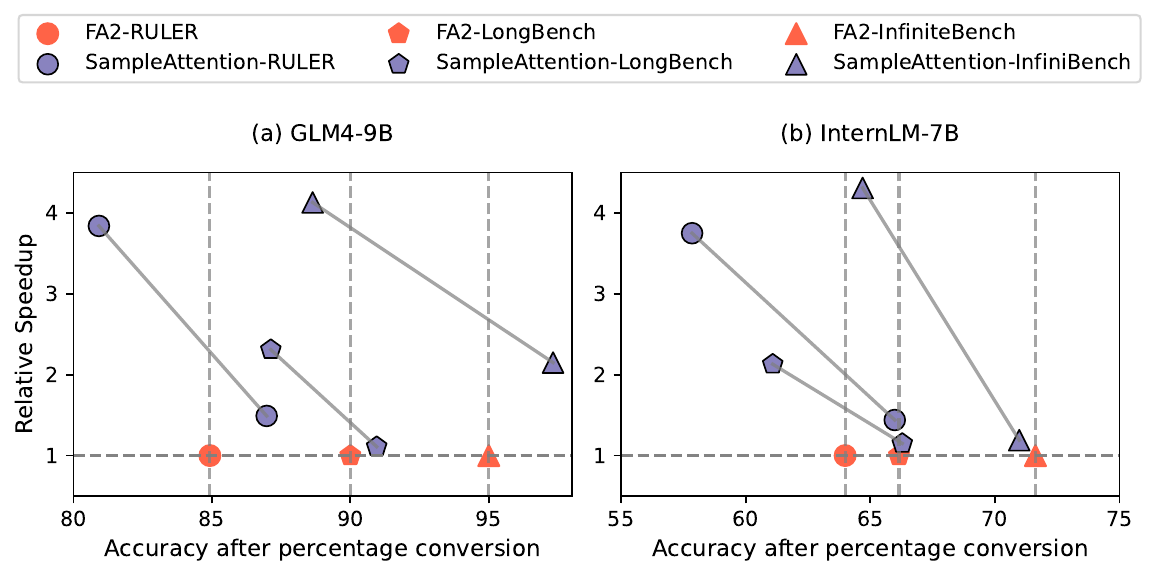}
  \caption{Results from offline tuning and evaluation of (a) GLM4-9B and (b) InternLM2-7B across RULER, LongBench, and InfiniteBench benchmarks. Different tasks share the same hyperparameters from offline tuning when sequence lengths fall within the same range.} 
  \label{robust}
\end{figure}
 
\textbf{\textbf{CRA} threshold $\alpha_{c}$ and $\alpha_{s}$:}
Tuning \(\alpha_c\) and \(\alpha_s\) within a given model is crucial for finding a cost-effective configuration. Generally, as shown in Figures~\ref{threshold-result} (a) and (c), increasing either \(\alpha_c\) or \(\alpha_s\) alone can improve accuracy, but at the expense of increased computational load. However, there are differences between them. The YI model in Figure~(c) at 128K shows significantly greater sensitivity to changes in the slash threshold \(\alpha_s\), whereas Figure~(a) demonstrates a more balanced response to changes in both thresholds.
Additionally, Figure~(b) shows that smaller thresholds can still deliver sufficiently good scores, enabling higher speedup while maintaining accuracy.

\textbf{Number of Sampling Chunks:} 
We further evaluate the impact of different numbers (1, 2, 4, 6) of sampling chunks on accuracy, given a fixed threshold. Table~\ref{chunk-study} demonstrates that an appropriate \(chunk_{n}\) size can yield more cost-effective results. For instance, increasing the number from 1 to 2 helps \WORK{} enhance accuracy without significantly impacting the speedup ratio. However, an excessively large number of chunks may not improve accuracy and can decrease the speedup. Thus, selecting an appropriate \(chunk_{n}\) size is essential.

\textbf{Cross-Task Robustness} 
The hyperparameters of SampleAttention are inherently model-specific due to architectural design and intrinsic sparsity patterns. To evaluate cross-task robustness, we tested shared hyperparameter configurations across three distinct benchmarks under the same model.
First, we performed tuning on GLM4 and InternLM2 using subsets of the RULER benchmark at various sequence lengths. We selected optimal configurations under two criteria: accuracy-optimized settings with negligible performance loss compared to full-attention baselines, and configurations maximizing acceleration gains. Detailed hyperparameters are provided in Appendix~\ref{hyper-para}.
Experimental results in Figure~\ref{robust} demonstrate that hyperparameters tuned on a subset of tasks generalize effectively across diverse benchmarks. For instance, GLM4's accuracy-optimized hyperparameters maintained near-lossless performance on different tasks from LongBench and InfiniteBench. This indicates robust cross-domain adaptability without significant performance degradation. Additionally, consistent speedup gains were observed for the same model under identical sequence lengths across tasks.

\begin{table}[ht]
  \caption{The impact of changing \(chunk_{n}\) on \textit{scores/speedup} in different cases. The scores above are based on RULER, while the speedup below are relative to FlashAttention2. The best score results are highlighted in \textbf{bold}, while the best speedup results are marked with \underline{\textbf{underline}}.}
  \label{chunk-study}
  \begin{adjustbox}{max width=0.8\textwidth, left}
    \renewcommand{\arraystretch}{0.9}
    \small
    \begin{tabular}{c|c|c|c|c|c}
      \toprule
      \multirow{2}{*}{\makecell[c]{Model}} & \multirow{2}{*}{($\alpha_{c}$,$\alpha_{s}$)} & \multicolumn{4}{c}{\(chunk_{n}\)}   \\
      \cmidrule(lr){3-6}
      & & 1 & 2 & 4 & 6 \\
      \midrule
      \multirow{2}{*}{\makecell[c]{GLM \\ (128K)}} & (0.90,0.90) & \makecell[c]{82.89/ \\ \underline{\textbf{1.92}}} & \makecell[c]{\textbf{84.70}/ \\1.89} & \makecell[c]{84.02/ \\1.70} & \makecell[c]{83.62/ \\1.53}\\ 
      &  (0.95,0.95) &   \makecell[c]{ 84.17/ \\1.64} & \makecell[c]{84.04/ \\1.60} &  \makecell[c]{83.14/ \\1.46} &  \makecell[c]{83.73/ \\1.33}\\
      \midrule
      \multirow{2}{*}{\makecell[c]{Yi \\(128K)}} & (0.95,0.95) & \makecell[c]{52.81/ \\ \underline{\textbf{2.17}}} & \makecell[c]{54.54/ \\ 2.12} & \makecell[c]{53.10/ \\1.89} &  \makecell[c]{54.58/ \\ 1.71} \\ 
      &  (0.98,0.98) & \makecell[c]{56.24/ \\ 1.29} & \makecell[c]{58.58/\\ 1.25} &  \makecell[c]{\textbf{59.37}/ \\1.21} & \makecell[c]{59.36/ \\ 1.13}\\
      
      \bottomrule
    \end{tabular}
  \end{adjustbox}
\end{table}

\subsection{Acceleration Speedup Benchmarking}

We conducted micro-benchmarks on a single A100 to evaluate the time breakdown and \texttt{TTFT} metrics. The baseline selected is FlashAttention2. All tests were conducted using the configuration from ChatGLM4-9B: 32 heads, and \(d = 128\), with synthetic data from the RULER benchmark as input. We standardized the batch size of the input data to 1 to support longer sequence lengths.
The hyperparameters are tuned to attain the best possible speedup without accuracy loss compared to FlashAttention2, ensuring the optimal acceleration-accuracy trade-off.
% The hyperparameters for each length are selected from tuning results to ensure an optimal speedup configuration with near-lossless accuracy compared to FlashAttention2. 
For example, the hyperparameters for 32K are set to \(\alpha_c = 0.95\), \(\alpha_s = 0.95\), and \(chunk_{n} = 1\). For lengths above 128K, the parameters tuned for 128K are reused. 
\begin{figure}[h]
  \centering
  \includegraphics[width=1.00\linewidth]{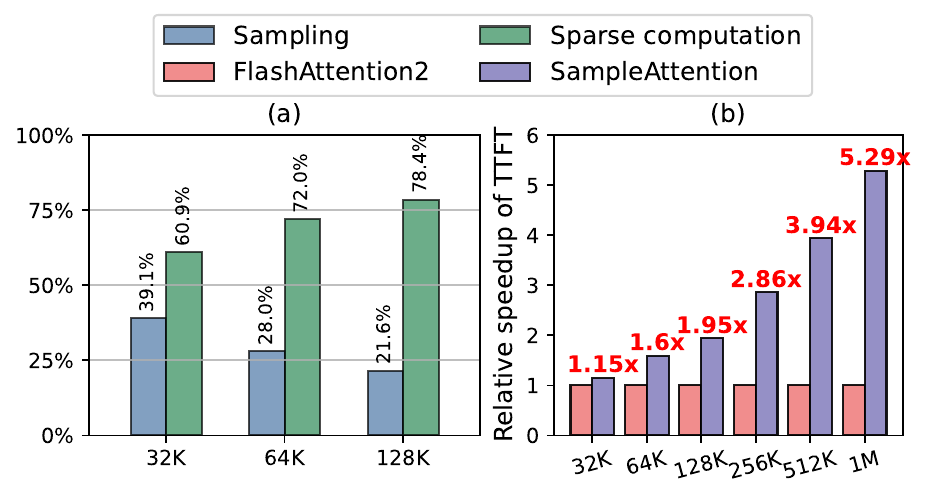}
  \caption{(a) The percentage  of time spent on sampling and sparse computation in \WORK{}. (b) Comparison of the TTFT metric using FlashAttention2 as the baseline. } 
  \label{benchmarking}
\end{figure}

\textbf{Sampling overhead.}
Figure~\ref{benchmarking}(a) presents the time breakdown for a full 40-layer model using synthetic data spanning sequence lengths from 32K to 128K. The results reveal that as sequence lengths increase, the relative contribution of sampling overhead diminishes. This trend underscores the potential of \WORK{} to deliver substantial acceleration advantages for longer sequences. However, in short-sequence scenarios, the performance gains are less pronounced due to the computational overhead associated with dynamic sampling and index construction. For instance, in the GLM4-9B model at an 8K sequence length, the latency of \WORK{} remains nearly identical to FlashAttention2 without compromising accuracy.

\textbf{Scaling the sequence length to 1M.}
We also conducted GPU performance evaluations scalable to a sequence length of 1 million.
To avoid memory issues, for sequence lengths greater than 256K, we divide the input sequence into multiple chunks and partition them by head during attention. We also further reduced a large number of intermediate results through operator fusion.
Figure~\ref{benchmarking}(b) shows that when the sequence length scales to 1M, the TTFT metric can be significantly reduced by \(5.29\times\).

\section{Related Work}
\label{sec:rel}

\textbf{Approximate Attention.} Plenty of works have been proposed to approximate quadratic attention with lower complexity\cite{ainslie2020etc, beltagy2020longformer, zaheer2020big, kitaev2020reformer, ding2023longnet, child2019generating, pagliardini2023faster, roy2021efficient, han2023hyperattention, wang2020linformer, choromanski2020rethinking, katharopoulos2020transformers, chen2021scatterbrain,chen2021pixelated, zhu2023biformer,ribar2023sparq, roy2021efficient, liu2023deja}. For example, BigBird~\cite{zaheer2020big} combines window-, global- and random-attention to capture long range dependency. Reformer~\cite{kitaev2020reformer} reduces computional cost via locality-sensitive hashing. LongNet~\cite{ding2023longnet} replaces full attention with dilated attention. Linformer~\cite{wang2020linformer} employs low-rank matrix to approximate attention. HyperAttention~\cite{han2023hyperattention} utilizes locality sensitive hashing to identify important entries on attention map. However, these approaches uses either static or coarse-grained sparse pattern, and often overlook the head-specific sparsity pattern. They cannot be losslessly applied in pretrained LLMs without additional finetuning or training. 
Additionally, although the recent work Minference~\cite{jiang2024minference} addresses the varying sparsity among attention heads by employing three distinct sparse attention patterns for long sequences, it relies on predefined fixed ratios for these sparse patterns and fails to account for the dynamic nature of the actual model and its prompts.

\textbf{KV Cache Compression.} Long sequence comes with substantial KV cache memory consumption. StreamingLLM~\cite{xiao2023efficient} keeps attention sinks and several recent tokens for infinite length generation. H2O~\cite{zhang2024h2o} dynamically retains a balance of recent and heavy hitter tokens according to attention score during decoding. FastGen~\cite{ge2023model} adaptively construct KV cache according to observed head-specific policies. SnapKV~\cite{li2024snapkv} strategically compresses the KV cache by selecting clustered critical positions for each attention head, leading to improvements in memory and latency efficiency during decoding. CHAI~\cite{agarwal2024chai} exemplifies head pruning methods that target reducing the KV cache overhead at the attention head level, thereby accelerating decoding.
Recent efforts also quantize KV cache to lower precision to reduce memory consumption~\cite{duanmu2024skvq,xiao2023smoothquant,zhao2023atom}. These works target on reducing the memory consumption of KV cache, while \WORK{} focuses on mitigating the long context computation overhead. \WORK{} can be combined with these approaches to further reduce memory consumption of KV cache.

% \red{
% \textbf{Distributed Mechanism for Long-Context Inference.} 
% To reduce the time-to-first-token and accelerate the inference of long-context sequences, leveraging the parallelism of multiple GPUs is also essential.
% Chunked pre-filling is a commonly adopted strategy~\cite{kwon2023efficient,agrawal2023sarathi,},  which involves dividing the prompt into fixed-length chunks to populate the KV cache beforehand. This practice substantially reduces peak memory usage by decreasing the peak intermediate activation size in linear layers from the sequence length to the chunk size.
% Additionally, the prefill-decode disaggregation technique~\cite{patel2024splitwise,zhong2024distserve} has been proposed to enhance the efficiency of the prefill stage by allocating sufficient resources to it. 
% Based on these mechanisms, Mooncake~\cite{qin2024mooncake} utilizes a separate node pool and employs chunked pipeline parallel (CPP) method to scale the processing of a single long-context request across multiple nodes. Compared to traditional sequence parallel approaches, CPP reduces network consumption and simplifies the reliance on frequent elastic scaling.
% }

\section{Conclusion}
In this paper, we identify the challenge of effectively exploiting the inherent high attention sparsity to accelerate prefill attention, due to the highly dynamic structured patterns and optimal sparsity ratios exhibited by the attention mechanism over long contexts. 
To address this, we propose \WORK{}, which utilizes \textbf{CRA} as a robust inficator of model accuracy and adaptively determines the sparsity ratio and pattern at runtime. 
Through an innovative two-stage query-guided filtering approach and hyperparameter tuning, it dynamically selects the minimal set of critical key-values, maximizing efficiency while maintaining accuracy. Experimental results demonstrate that \WORK{} consistently maintains robust accuracy across various benchmarks, models, and sequence lengths, and significantly reduces the TTFT metric.

% Acknowledgements should only appear in the accepted version.

% \nocite{langley00}
\nocite{*}
\bibliographystyle{mlsys2025} 
\bibliography{example_paper}
\clearpage
\appendix
%%%%%%%%%%%%%%%%%%%%%%%%%%%%%%%%%%%%%%%%%%%%%%%%%%%%%%%%%%%%
\newpage
\section{Appendix}

\subsection{Visualization of Attention Score}
\label{app:vis_attn}

\begin{figure*}[htp]
    \centering
    \begin{subfigure}[b]{0.25\textwidth}
        \centering
        \includegraphics[width=1\textwidth]{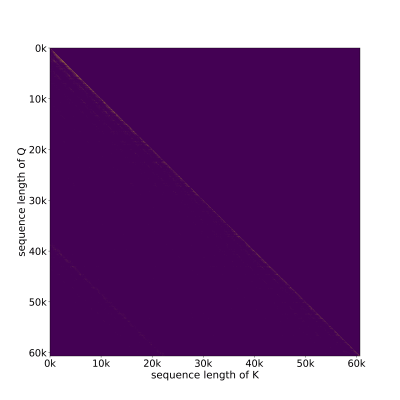}
        \caption{Layer0}
    \end{subfigure}%
    \begin{subfigure}[b]{0.25\textwidth}
        \centering
        \includegraphics[width=1\textwidth]{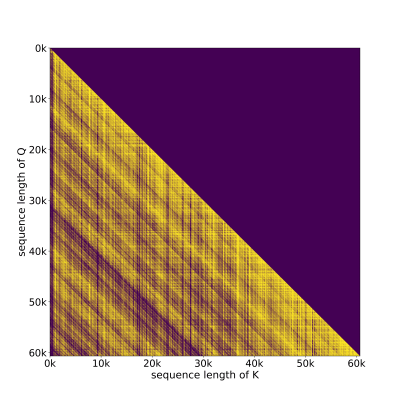}
        \caption{Layer0}
    \end{subfigure}%
    \begin{subfigure}[b]{0.25\textwidth}
        \centering
        \includegraphics[width=1\textwidth]{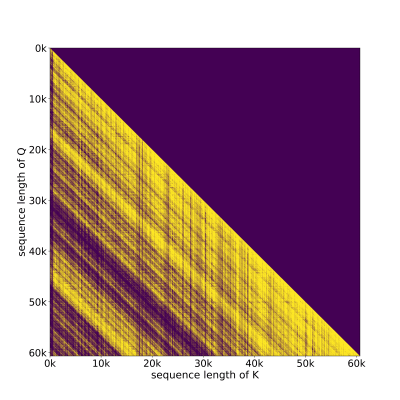}
        \caption{Layer0}
    \end{subfigure}%
    \begin{subfigure}[b]{0.25\textwidth}
        \centering
        \includegraphics[width=1\textwidth]{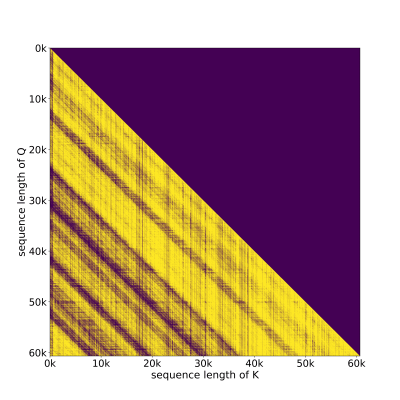}
        \caption{Layer0}
    \end{subfigure}

    \begin{subfigure}[b]{0.25\textwidth}
        \centering
        \includegraphics[width=1\textwidth]{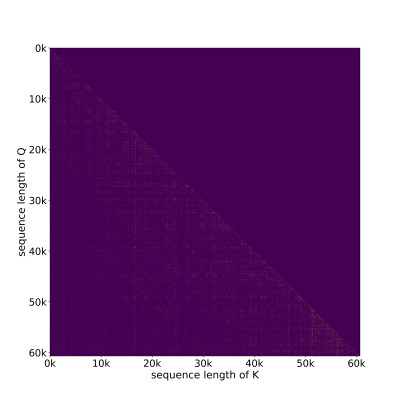}
        \caption{Layer4}
    \end{subfigure}%
    \begin{subfigure}[b]{0.25\textwidth}
        \centering
        \includegraphics[width=1\textwidth]{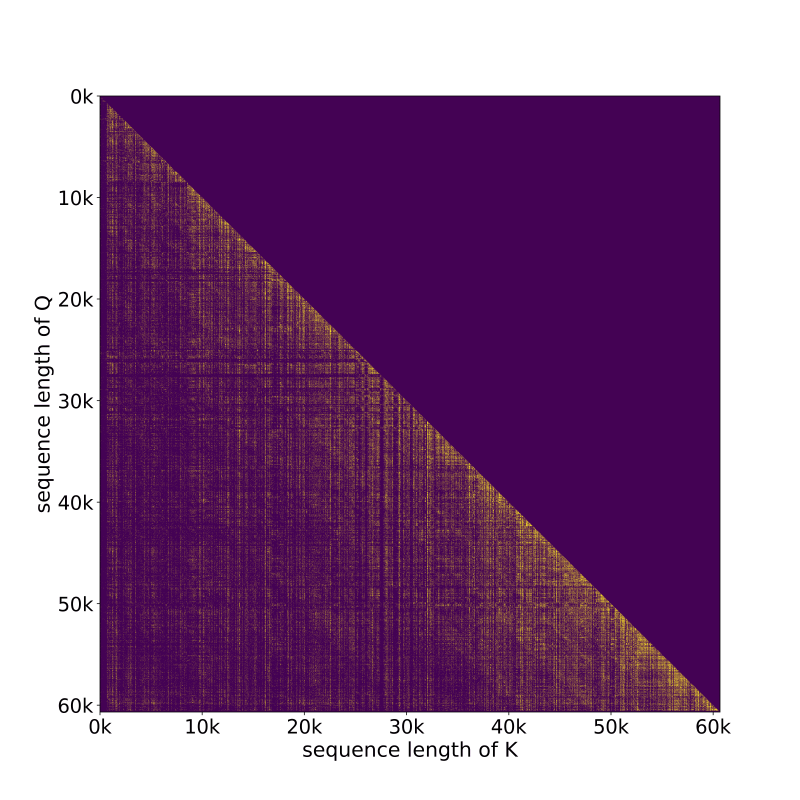}
        \caption{Layer4}
    \end{subfigure}%
    \begin{subfigure}[b]{0.25\textwidth}
        \centering
        \includegraphics[width=1\textwidth]{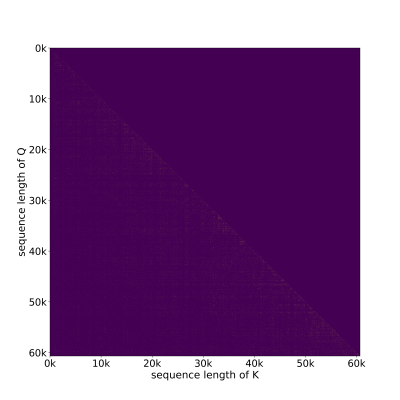}
        \caption{Layer4}
    \end{subfigure}%
    \begin{subfigure}[b]{0.25\textwidth}
        \centering
        \includegraphics[width=1\textwidth]{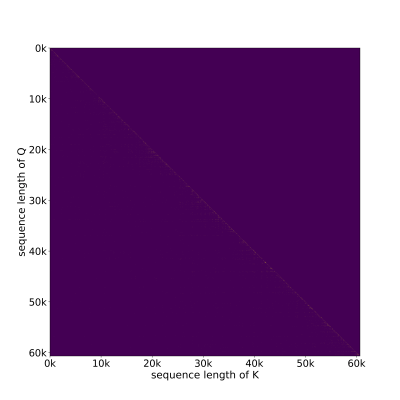}
        \caption{Layer4}
        
    \end{subfigure}
    
    \begin{subfigure}[b]{0.25\textwidth}
        \centering
        \includegraphics[width=1\textwidth]{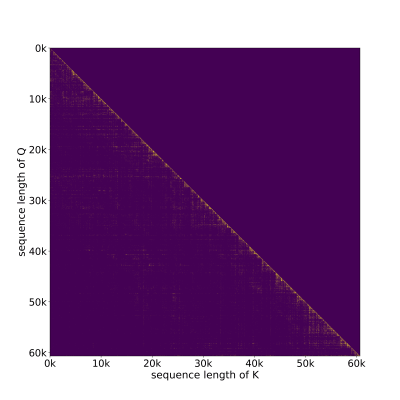}
        \caption{Layer8}
    \end{subfigure}%
    \begin{subfigure}[b]{0.25\textwidth}
        \centering
        \includegraphics[width=1\textwidth]{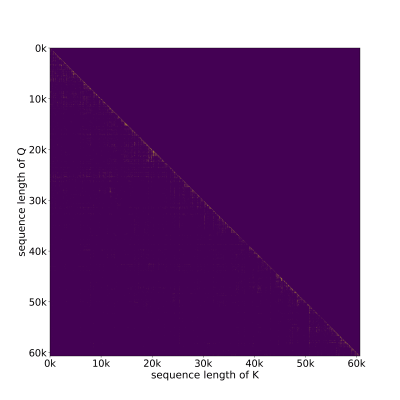}
        \caption{Layer8}
    \end{subfigure}%
    \begin{subfigure}[b]{0.25\textwidth}
        \centering
        \includegraphics[width=1\textwidth]{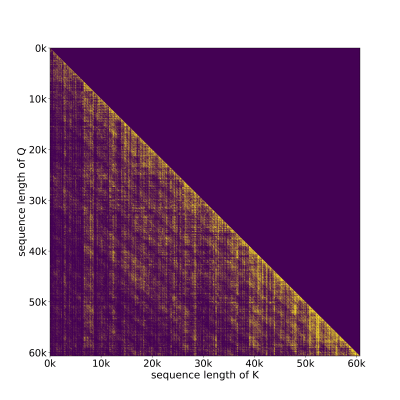}
        \caption{Layer8}
    \end{subfigure}%
    \begin{subfigure}[b]{0.25\textwidth}
        \centering
        \includegraphics[width=1\textwidth]{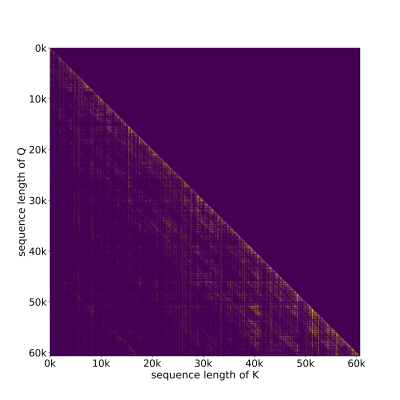}
        \caption{Layer8}
    \end{subfigure}

    \begin{subfigure}[b]{0.25\textwidth}
        \centering
        \includegraphics[width=1\textwidth]{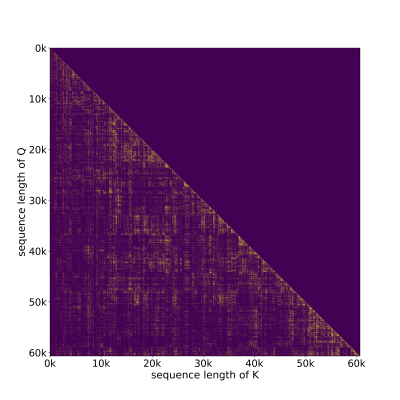}
        \caption{Layer12}
    \end{subfigure}%
    \begin{subfigure}[b]{0.25\textwidth}
        \centering
        \includegraphics[width=1\textwidth]{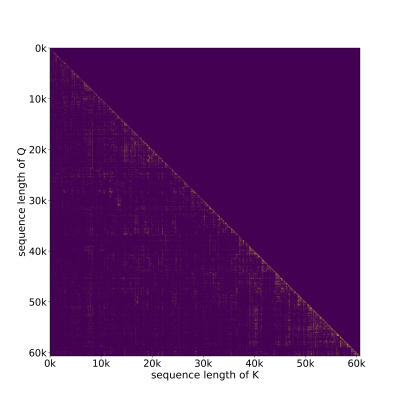}
        \caption{Layer12}
    \end{subfigure}%
    \begin{subfigure}[b]{0.25\textwidth}
        \centering
        \includegraphics[width=1\textwidth]{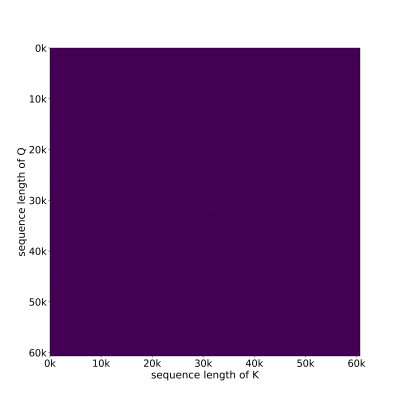}
        \caption{Layer12}
    \end{subfigure}%
    \begin{subfigure}[b]{0.25\textwidth}
        \centering
        \includegraphics[width=1\textwidth]{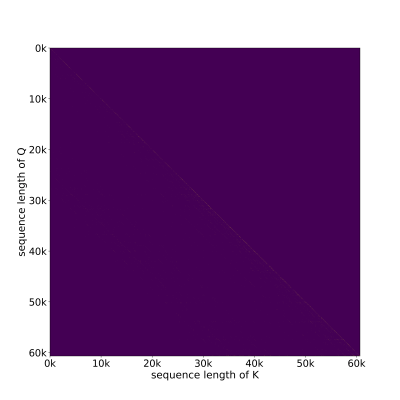}
        \caption{Layer12}
    \end{subfigure}
    \caption{The visualization attention based on a content length of 61K, displays the sparse patterns for randomly chosen heads from layers 0, 4, 8 and 12.}
    \label{visualization1}
\end{figure*}

\begin{figure*}[htp]
    \centering
     
    \begin{subfigure}[b]{0.25\textwidth}
        \centering
        \includegraphics[width=1\textwidth]{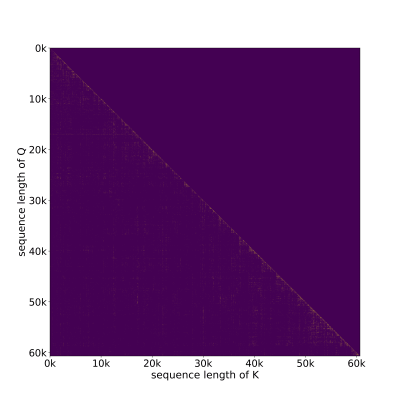}
        \caption{Layer16}
    \end{subfigure}%
    \begin{subfigure}[b]{0.25\textwidth}
        \centering
        \includegraphics[width=1\textwidth]{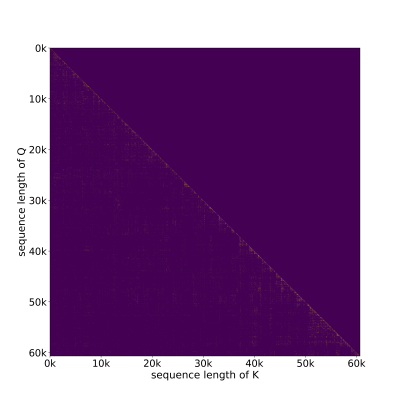}
        \caption{Layer16}
    \end{subfigure}%
    \begin{subfigure}[b]{0.25\textwidth}
        \centering
        \includegraphics[width=1\textwidth]{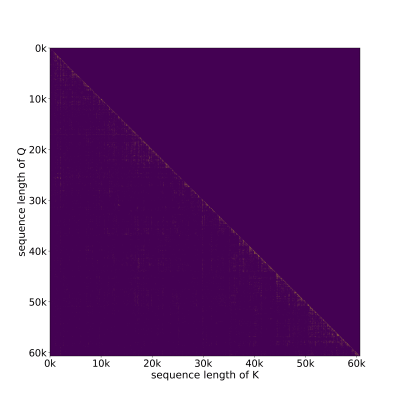}
        \caption{Layer16}
    \end{subfigure}%
    \begin{subfigure}[b]{0.25\textwidth}
        \centering
        \includegraphics[width=1\textwidth]{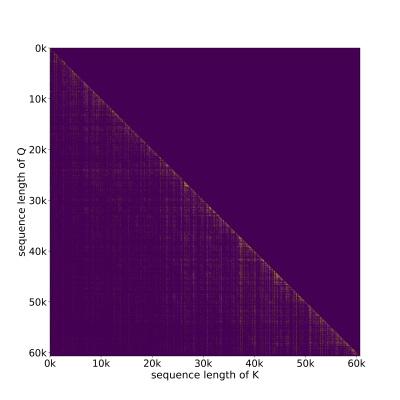}
        \caption{Layer16}
    \end{subfigure}

     \begin{subfigure}[b]{0.25\textwidth}
        \centering
        \includegraphics[width=1\textwidth]{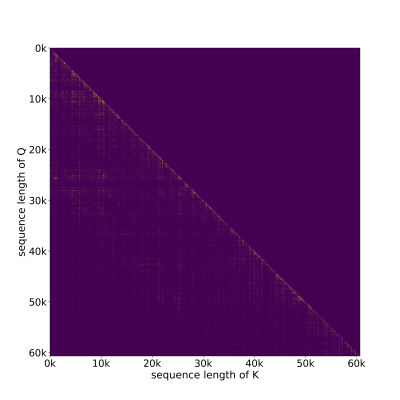}
        \caption{Layer20}
    \end{subfigure}%
    \begin{subfigure}[b]{0.25\textwidth}
        \centering
        \includegraphics[width=1\textwidth]{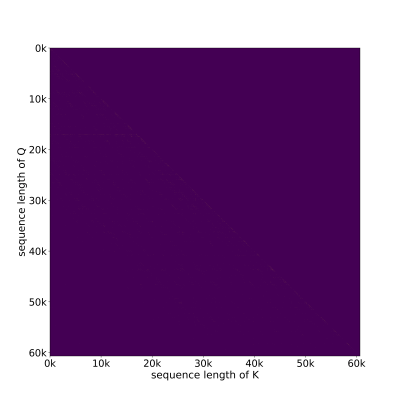}
        \caption{Layer20}
    \end{subfigure}%
    \begin{subfigure}[b]{0.25\textwidth}
        \centering
        \includegraphics[width=1\textwidth]{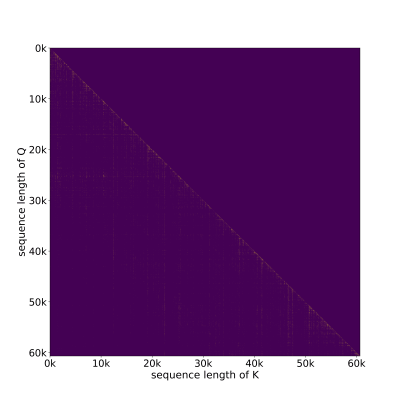}
        \caption{Layer20}
    \end{subfigure}%
    \begin{subfigure}[b]{0.25\textwidth}
        \centering
        \includegraphics[width=1\textwidth]{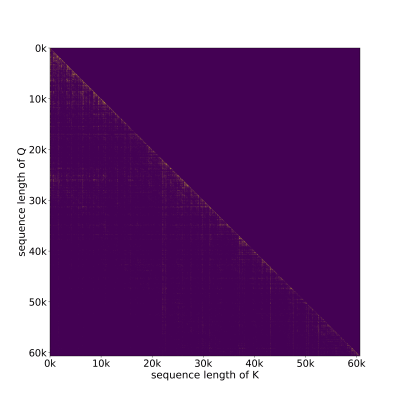}
        \caption{Layer20}
    \end{subfigure}

    \begin{subfigure}[b]{0.25\textwidth}
        \centering
        \includegraphics[width=1\textwidth]{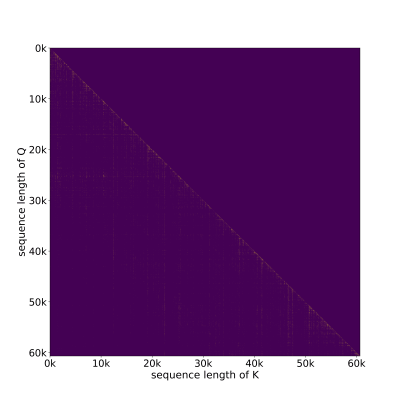}
        \caption{Layer24}
    \end{subfigure}%
    \begin{subfigure}[b]{0.25\textwidth}
        \centering
        \includegraphics[width=1\textwidth]{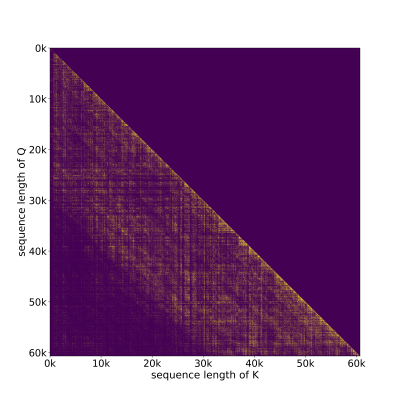}
        \caption{Layer24}
    \end{subfigure}%
    \begin{subfigure}[b]{0.25\textwidth}
        \centering
        \includegraphics[width=1\textwidth]{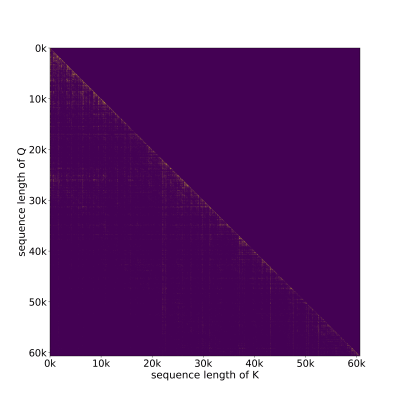}
        \caption{Layer24}
    \end{subfigure}%
    \begin{subfigure}[b]{0.25\textwidth}
        \centering
        \includegraphics[width=1\textwidth]{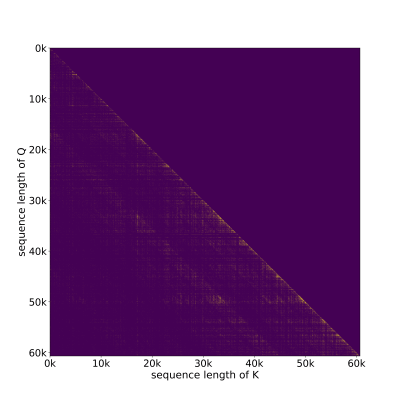}
        \caption{Layer24}
    \end{subfigure}
    \caption{The visualization attention based on a content length of 61K, displays the sparse patterns for randomly chosen heads from layers 16, 20 and 24.}
    \label{visualization2}
\end{figure*}

Figures~\ref{visualization1} and Figures~\ref{visualization2} present the sparse patterns across various heads in the ChatGLM3-6B model (28 layers x 32 heads) under a sequence length of 61K. We conducted row-by-row filtering based on the full attention softmax weight, using a \textbf{CRA} threshold of $\alpha=0.95$, and randomly selected four heads from different layers for display.

According to the visualization results on the majority of heads, we observed two distinct and prominent patterns prevalent in the heatmap of attention weight: column stripes and slash stripes. Column stripe patterns embody the global contextual information whereas slash stripe capture local information. 

% \clearpage

\subsection{Sparisty Analysis}
\label{appendix-sparsity}

To further quantify the degree of sparsity exposed as sequence lengths increase, we conducted scalability tests on the ChatGLM3-6B model using the "Needle-in-a-Haystack" task to evaluate sparsity. The results are presented in Table~\ref{analysis-sparsity}. 
According to the results, the increase in sequence length introduces more apparent sparsity.
With each doubling of length, the proportion of KV elements needed to maintain the same threshold $\alpha$ decreases by approximately 20\%.
Concurrently, a smaller threshold results in the filtering of more KV elements, which may also lead to a decline in task performance in  accuracy. 

\begin{table}[ht]
  \caption{Sparsity analysis for the "Needle in a Haystack" task}
  \label{analysis-sparsity}
  \centering
  \begin{adjustbox}{max width=\textwidth, left}
  \renewcommand{\arraystretch}{0.95}
  \small
  \begin{tabular}{c|c|c}
    \toprule
    % \multicolumn{2}{c}{Part}                   \\
    \cmidrule(r){1-2}
    \makecell[c]{Sequence Length}     &  \makecell[c]{Average  Sparsity\\ in ChatGLM-6B }   &  \makecell[c]{Average  Sparsity \\ in InternLM-7B} \\
    \midrule
    4K &   88.00\%  &    91.13\%   \\
    8K &   90.74\%  &    92.72\%    \\
    16K  &   92.52\%  &    93.89\%    \\
    32K  &   93.88\%  &    94.83\%     \\
    64K  &   94.89\%  &    95.89\%     \\
    128K  &   95.84\%  &    96.67\%     \\
    \bottomrule
  \end{tabular}
  \end{adjustbox}
\end{table}

\subsection{Effectiveness of sampling}
\label{appendix-sampling}

To validate the efficiency of this chunked sampling method, we conducted tests using two different sampling ratios on three different heads. One sampling rate was set to full sampling, while the other was configured with a query block size of 128 and a sampling rate where $chunk_{n}$ was 2. We applied varying proportions of \texttt{top-k} columns and slash stripes to the full attention matrix as masks to observe the changes in \textbf{CRA}. The results, shown in Table~\ref{sampling-study}, indicate that the CRA obtained by selecting \texttt{top-k} stripes at a 0.4\% sampling rate is very close to the CRA obtained from the complete attention scores, with the difference decreasing as the proportion increases. This demonstrates that the chunked sampling method in \WORK{} is both simple and efficient.

\begin{table*}[ht]
  \caption{
  The \textbf{CRA} percentages can be achieved by selecting different ratios of \textit{top-k} stripes under varying sampling rates. The tests were conducted on the RULER task with a sequence length of 64K using the GLM4-9B model.} 
  % \JF{Mark the second best with underline, e.g. \underline{158.75}}
  \label{sampling-study}
  \begin{adjustbox}{max width=\textwidth, left}
    \begin{tabular}{c|c|c|c|c|c|c|c|c|c|c}
      \toprule
      {\makecell[c]{ratio of \\ \texttt{top-k}}} &  \multicolumn{2}{c|}{2.5\%} &\multicolumn{2}{c|}{10\%} &\multicolumn{2}{c|}{20\%} 
 &\multicolumn{2}{c|}{40\%} & \multicolumn{2}{c}{80\%} \\ 
     \cmidrule(lr){1-1}
      \cmidrule(lr){2-3}
      \cmidrule(lr){4-5}
      \cmidrule(lr){6-7}
      \cmidrule(lr){8-9}
      \cmidrule(lr){10-11}
 sampling ratio & 100\% & 0.4\% & 100\% & 0.4\% & 100\% & 0.4\% & 100\% & 0.4\% & 100\% & 0.4\%  \\
\midrule
    \makecell[c]{HEAD-1} & 16.35\% &12.74\% & 26.91\% 
& 23.33\%& 45.99\%& 42.14\% &58.21\% &55.34\% & 96.30\% & 93.65\% 
  \\
    \makecell[c]{HEAD-2}   & 55.43\% &48.40\% & 63.89\% & 58.63\% & 85.92\% &81.98\% &89.07\% &84.21\% &99.15\% &98.08\%     \\
    \makecell[c]{HEAD-3}  
&93.20\%&90.44\%&98.32\%&97.62\%&99.14\%&98.43\%&99.41\%&99.12\%&99.98\%&99.66\%  \\
      \bottomrule
    \end{tabular}
  \end{adjustbox}
\end{table*}

\begin{table*}[ht]
  \caption{Hyperparameters selected for GLM-4B and InterLM2-7B after tuning across sequence length ranges. Acc | Spd represents the optimal configuration that maintains accuracy without loss compared to the full-attention, as well as the configuration achieving the highest speedups while retaining at least 90\% of the  accuracy.} 
  \label{hyper-para}
  \begin{adjustbox}{max width=\textwidth, left}
    \begin{tabular}{c|c|c|c|c|c|c|c|c|c|c}
      \toprule
      {\makecell[c]{range of \\ sequence length}} &  \multicolumn{2}{c|}{ < 16K } &\multicolumn{2}{c|}{[16K,48K)} &\multicolumn{2}{c|}{[48K,80K)} 
 &\multicolumn{2}{c|}{[80K,112K)} & \multicolumn{2}{c}{>=112K} \\ 
     \cmidrule(lr){1-1}
      \cmidrule(lr){2-3}
      \cmidrule(lr){4-5}
      \cmidrule(lr){6-7}
      \cmidrule(lr){8-9}
      \cmidrule(lr){10-11}
  {\makecell[c]{models \\ Acc | Spd}}  & GLM4& InternLM2 & GLM4& InternLM2 & GLM4& InternLM2 & GLM4& InternLM2 & GLM4& InternLM2  \\
\midrule
    \makecell[c]{CRA \\ Column} & 0.98 | 0.85 & 0.98 | 0.85 & 0.95 | 0.85 & 0.95 | 0.80      & 0.95 | 0.80 & 0.92 | 0.80     & 0.90 | 0.80 & 0.92 | 0.80      & 0.95 | 0.80 & 0.95 | 0.80
  \\
    \makecell[c]{CRA \\ Slash}   & 0.90 | 0.85  & 0.95 | 0.85 &  0.95 | 0.80  & 0.90 | 0.80    & 0.95 | 0.80  & 0.92 | 0.80 & 0.90 | 0.80 & 0.90 | 0.80 & 0.85 | 0.80 & 0.90 | 0.80   \\
    \makecell[c]{Num of \\ Chunks}  
&1 | 1 & 1 | 1 & 1 | 1 & 1 | 1&  1 | 1 & 1 | 1& 2 | 1 & 2 | 1& 1 | 1  & 2 | 1 \\
      \bottomrule
    \end{tabular}
  \end{adjustbox}
\end{table*}

\end{document}